\pdfoutput=1
\documentclass{article}

\usepackage{microtype}
\usepackage{graphicx}

\usepackage{booktabs} 

\usepackage{hyperref}

\usepackage[accepted]{icml2025}
\usepackage{subcaption}
\usepackage{pifont}

\usepackage[skins,breakable]{tcolorbox}
\usepackage{xcolor}

\newtcolorbox{promptbox}[1][]{
    enhanced,
    breakable,
    colback=gray!5!white,      
    colframe=gray!50!black,    
    boxrule=0.5pt,             
    arc=3pt,                   
    left=6pt, right=6pt, top=6pt, bottom=6pt,
    fonttitle=\bfseries\sffamily,
    coltitle=black,
    attach boxed title to top left={yshift=-2mm, xshift=4mm},
    boxed title style={colback=gray!15!white, colframe=gray!50!black, boxrule=0.5pt, arc=2pt},
    title=#1
}

\usepackage{pifont}

\newcommand{\cmark}{\ding{51}} 
\newcommand{\xmark}{\ding{55}} 

\usepackage[table]{xcolor}
\usepackage{amsmath}
\usepackage{amssymb}
\usepackage{mathtools}
\usepackage{amsthm}

\usepackage[capitalize,noabbrev]{cleveref}

\usepackage{booktabs} 
\usepackage{multirow} 
\usepackage{graphicx} 
\usepackage{caption}  

\usepackage{listings}
\usepackage{xcolor} 
\theoremstyle{plain}

\theoremstyle{definition}

\theoremstyle{remark}

\usepackage[textsize=tiny]{todonotes}

\icmltitlerunning{GeoSeg: Training-Free Reasoning-Driven Segmentation in Remote Sensing Imagery}

\begin{document}

\twocolumn[
\icmltitle{GeoSeg: Training-Free Reasoning-Driven Segmentation
\\ in Remote Sensing Imagery}



\icmlsetsymbol{equal}{*}

\begin{icmlauthorlist}
\icmlauthor{Lifan Jiang}{sch}
\icmlauthor{Yuhang Pei}{sch}
\icmlauthor{Boxi Wu}{sch}
\icmlauthor{Yan Zhao}{cmp1}
\icmlauthor{Tianrun Wu}{sch}
\icmlauthor{Shulong Yu}{sch}
\icmlauthor{Lihui Zhang}{cmp2}
\icmlauthor{Deng Cai}{sch}

\end{icmlauthorlist}

\icmlaffiliation{sch}{State Key lab of CAD\&CG, Zhejiang University}
\icmlaffiliation{cmp1}{UniTTEC Co. Ltd.}
\icmlaffiliation{cmp2}{Wuchan Zhongda Chengtou (Ningbo) Holdings. Ltd.}

\icmlcorrespondingauthor{Lifan Jiang}{lifanjiang@zju.edu.cn}

\icmlkeywords{Machine Learning, ICML}

\vskip 0.3in

]




\printAffiliationsAndNotice{} 

\begin{abstract}
Recent advances in MLLMs are reframing segmentation from fixed-category prediction to instruction-grounded localization. While reasoning based segmentation has progressed rapidly in natural scenes, remote sensing lacks a generalizable solution due to the \textbf{prohibitive cost} of reasoning-oriented data and domain-specific challenges like overhead viewpoints. We present \textbf{GeoSeg}, a \textbf{zero-shot, training-free} framework that bypasses the supervision bottleneck for reasoning-driven remote sensing segmentation. GeoSeg couples MLLM reasoning with precise localization via: (i) \textbf{bias-aware coordinate refinement} to correct systematic grounding shifts and (ii) a \textbf{dual-route prompting} mechanism to fuse semantic intent with fine-grained spatial cues. We also introduce \textbf{GeoSeg-Bench}, a diagnostic benchmark of 810 image--query pairs with hierarchical difficulty levels. Experiments show that GeoSeg consistently outperforms all baselines, with extensive ablations confirming the effectiveness and necessity of each component.

\end{abstract}

\section{Introduction}
\label{sec:intro}

Segmentation is a cornerstone of visual understanding, yet its formulation evolves with how we specify \textit{what} to segment. In remote sensing, early work~\cite{deeplabv3,factseg,resunet,transunet,rsmamba} largely followed a closed-set paradigm with dense pixel supervision over a fixed label space. Subsequent open-vocabulary approaches~\cite{remoteclip,ODISE,OpenRSS,OVSeg,openseg,OWL-ViT} leveraged vision--language alignment to generalize beyond the training taxonomy. More recently, generalist promptable segmenters~\cite{sam,sam2,sam3,samrs} further decoupled segmentation from category inventories by taking points or boxes as guidance, making ``segment anything'' a practical primitive.

\begin{figure*}[t]
    \centering
    \includegraphics[width=0.94\textwidth]{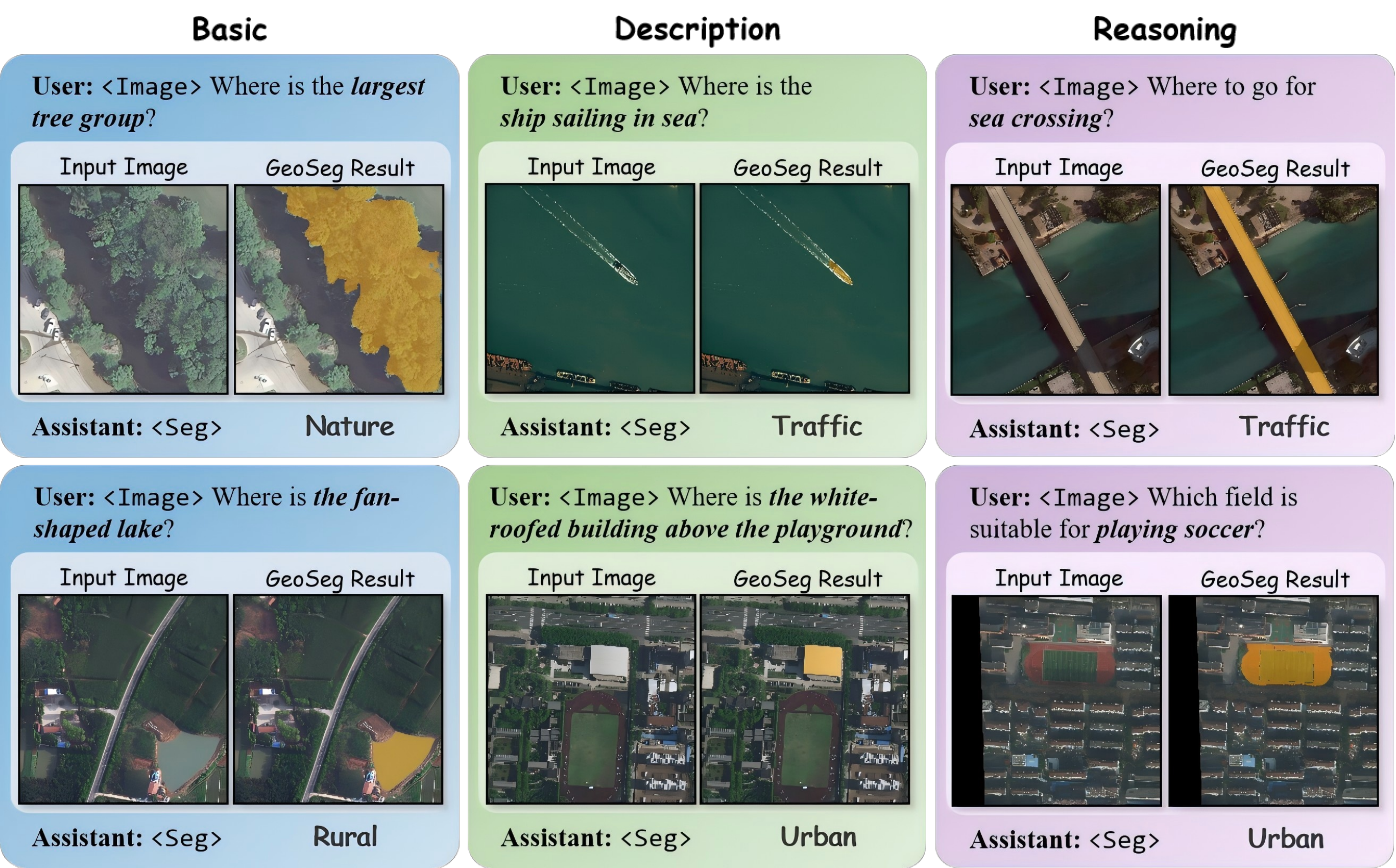} 
    \captionof{figure}{\textbf{Performance across reasoning difficulty levels.} 
    We visualize our proposed GeoSeg's results on image--query pairs across three difficulty levels. The visualized masks illustrate that GeoSeg can handle varying instruction complexity and remain effective in challenging scenarios. Please refer to Appendix~\ref{app:baseline} for more results.}
    \label{fig:visible}
\end{figure*}

A parallel and increasingly important direction is instruction and reasoning-driven segmentation~\cite{lisa,visa,lasagna,psalm,gsva}. Instead of being told a category name or a point prompt, the model must interpret a natural language request that can involve attributes, relations, and implicit intent (e.g., \textit{``the residential buildings arranged in rows next to the park''} or \textit{``where can I seek medical help in an emergency?''}), and then ground the answer into a pixel-level mask. In natural images, recent progress shows that combining multimodal language understanding with segmentation can support such reasoning-oriented queries. However, the extension of these advances to remote sensing is hindered by a structural domain gap. Modern MLLMs, conditioned on gravity-aligned natural scenes, often struggle with the rotation-invariant visual statistics of overhead imagery. This misalignment is consequential: without robust grounding, remote-sensing segmentation remains shackled to fixed taxonomies, limiting its utility for open-ended analysis.

Remote sensing scenes pose unique challenges that make reasoning segmentation particularly non-trivial. \textbf{First}, the overhead perspective changes how objects appear and removes many cues common in natural images. \textbf{Second}, drastic scale variations and high object density complicate localization: the same concept may occupy wildly different pixel footprints, while visually similar structures can be tightly packed. \textbf{Third}, remote sensing objects often exhibit weak texture differences and are better distinguished through spatial context (e.g., adjacency, layout, road connectivity) or functional semantics. \textbf{Finally}, while reasoning-based segmentation thrives on diverse, instruction-rich supervision, the remote sensing domain suffers from a scarcity of reasoning-oriented datasets, making heavy training and task-specific adaptation less attractive. Collectively, these factors leave a clear gap: despite the maturation of closed-set, open-vocabulary, and generalist segmentation, the remote sensing community still lacks a generalizable and \textbf{training-free} paradigm for reasoning-driven segmentation, as well as a dedicated benchmark to measure progress.

We introduce \textbf{GeoSeg}, a training-free framework tailored for reasoning-driven segmentation in remote sensing imagery. Our goal is to follow open-ended instructions without additional training, thereby circumventing the supervision bottleneck. GeoSeg couples the reasoning capability of multimodal large language models (MLLMs) with the precise localization of promptable segmenters, and makes this coupling reliable through two designs: (i) \textbf{bias-aware coordinate refinement} to correct systematic grounding shifts under overhead views, and (ii) \textbf{dual-route prompting} to fuse coarse semantic intent with fine-grained visual keypoints. Throughout the paper, we use \emph{MLLM} and \emph{VLM} interchangeably to denote multimodal foundation models with image--text inputs; we denote the reasoning/grounding model as $\mathcal{L}$ and the evaluation judge model as $\mathcal{J}$. This combination enables GeoSeg to handle diverse query forms---from explicit descriptions to implicit intent---while producing accurate pixel masks.

To enable rigorous evaluation and diagnosis, we further introduce GeoSeg-Bench, a dedicated benchmark comprising 810 image--query pairs spanning diverse scenarios with hierarchical difficulty levels. GeoSeg-Bench supports comprehensive comparison across general segmentation models, reasoning-oriented segmentation approaches, and modern MLLMs under a unified test-only protocol, i.e., no fine-tuning on GeoSeg-Bench. Experiments show that GeoSeg achieves the best overall performance on GeoSeg-Bench across standard pixel-level metrics while exhibiting highly competitive inference efficiency, and extensive ablations verify that each component is necessary for the final performance.

\begin{figure*}[t]
    \centering
    \setlength{\fboxsep}{0pt} 
    \includegraphics[width=\textwidth]{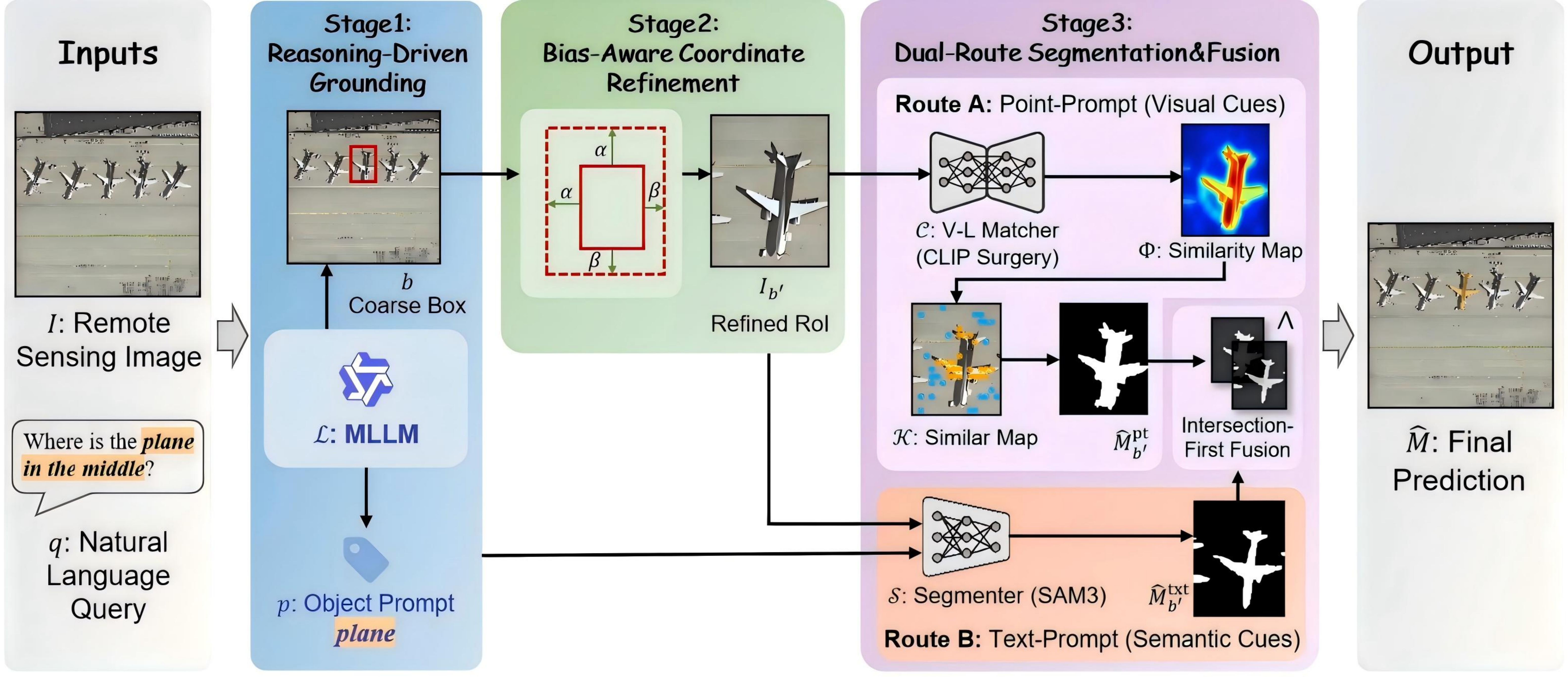}
    
    \caption{\textbf{Overview of the GeoSeg pipeline.} 
    Given a remote sensing image $I$ and a natural language query $q$, the pipeline operates in three stages:
    (1) \textbf{Reasoning-Driven Grounding}: The MLLM ($\mathcal{L}$) generates a coarse bounding box $b$ and extracts the object prompt $p$.
    (2) \textbf{Bias-Aware Coordinate Refinement}: To mitigate grounding bias, the box is adjusted via asymmetric expansion ($\alpha, \beta$) to yield a refined RoI $I_{b'}$.
    (3) \textbf{Dual-Route Segmentation \& Fusion}: Within the RoI, we perform parallel segmentation using \textit{Route A} (Visual Cues via CLIP Surgery) and \textit{Route B} (Semantic Cues via SAM3 with prompt $p$). 
    The final prediction $\hat{M}$ is obtained by integrating both paths via Intersection-First Fusion.}
    \label{fig:pipeline}

\end{figure*}

In summary, our contributions are three-fold:
\begin{itemize}
    \item \textbf{Task \& Problem Setting:} We introduce the problem setting of \emph{instruction-grounded, reasoning-driven} segmentation in remote sensing, and identify the key challenges that differentiate it from natural-image benchmarks.

    \item \textbf{Methodological Innovation:} We propose GeoSeg, a training-free framework that integrates bias-aware coordinate refinement and dual-route prompting to enable accurate instruction-grounded pixel-level localization.

    \item \textbf{Benchmark \& Evaluation:} We establish GeoSeg-Bench, a dedicated benchmark with 810 image--query pairs and hierarchical difficulty levels, and provide a standardized evaluation protocol that supports comprehensive comparison across reasoning-based segmentation methods.
\end{itemize}
\vspace{-8pt}

\noindent\textbf{Reproducibility:} Code, evaluation prompts, and GeoSeg-Bench will be publicly released. Additional qualitative examples are provided in Appendix ~\ref{app:baseline}.

\vspace{-3pt}
\section{Related Work}
\vspace{-3pt}
\label{sec:related}

\subsection{Remote Sensing Segmentation Evolution}

Remote sensing segmentation has long been dominated by the closed-set supervised paradigm. Traditional models~\cite{deeplabv3,factseg,resunet,transunet,rsmamba} predict pixel masks over a predefined taxonomy, relying on dense annotations to achieve strong performance, yet are bound by costly supervision and limited category coverage. To relax the fixed-label constraint, open-vocabulary segmentation~\cite{remoteclip,ODISE,OpenRSS,OVSeg,openseg,clipseg,groupvit,denseclip} leverages vision--language pretraining to generalize beyond the training taxonomy via image--text alignment. Despite this progress, most open-vocabulary methods in remote sensing still operate on explicit class names or short phrases; they struggle to interpret instruction-grounded queries involving complex spatial relations, fine-grained attributes, or implicit intent. This limitation motivates a paradigm shift from simple vocabulary expansion to reasoning-based localization under open-ended instructions.

\subsection{Generalist Promptable Models}

Foundation promptable segmenters~\cite{sam,sam2,sam3,samrs} have decoupled segmentation from category inventories by taking points or boxes as prompts, establishing ``segment anything'' as a practical primitive. This capability has catalyzed efforts to adapt such models to remote sensing. A common recipe is to pair a grounding/detection module~\cite{Gclip,grounddino} with a SAM-like segmenter, using predicted boxes or points as prompts. While effective when prompts are accurate, these pipelines typically treat language understanding as an external component and remain highly sensitive to grounding biases under overhead views, often leading to cascading segmentation failures. In contrast, our approach emphasizes robust grounding rectification and reliable prompt generation, ensuring accurate segmentation even without high-quality \textit{a priori} prompts.

\begin{figure}[t]
    \centering
    \includegraphics[width=0.48\textwidth]{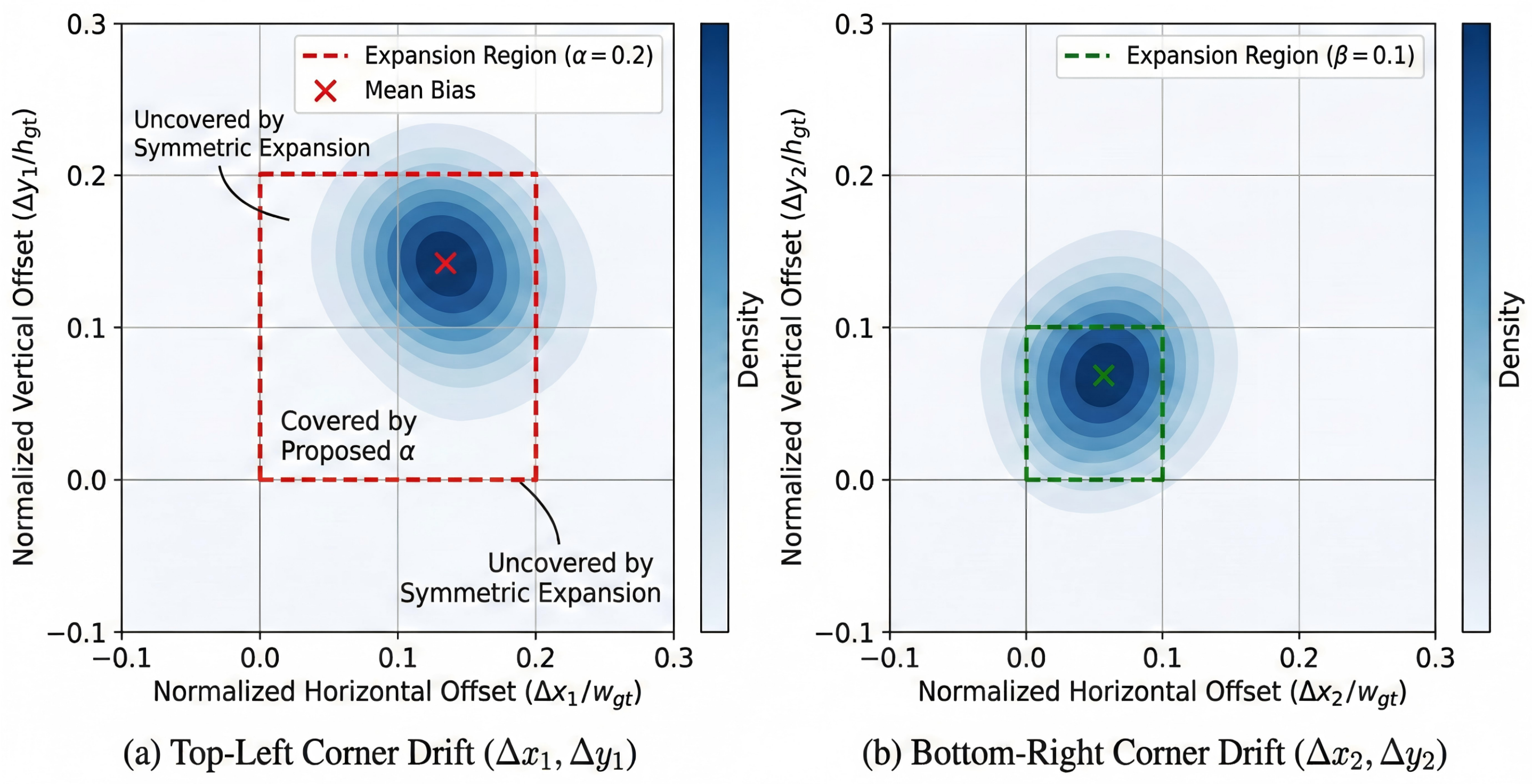}
    \caption{\textbf{Quantification of domain-specific grounding drift.} 
    We analyze coordinate offsets on a held-out calibration set comprising 1,000 images randomly sampled from LoveDA, NWPU-VHR-10, and DIOR datasets. 
    The KDE visualization reveals a systematic bottom-right shift inherent to pre-trained MLLMs under overhead views, necessitating our statistically derived asymmetric expansion ($\alpha=0.2, \beta=0.1$).}
    \label{fig:bbox_bias}
\end{figure}

\subsection{Reasoning-Driven Segmentation}
In natural images, language-guided segmentation~\cite{lisa,visa,lasagna,psalm,gsva,segzero,seem,x-decoder} has evolved from simple referring expressions to sophisticated reasoning-driven tasks, where models interpret complex queries to output pixel-level masks. However, existing approaches are predominantly tailored for natural scenes and typically necessitate extensive fine-tuning on instruction--mask datasets, which are substantially scarcer in remote sensing. Furthermore, the significant domain gap---characterized by overhead viewpoints, extreme scale variations, and context-dependent functional semantics---hinders direct transfer. Distinct from prior work, we explore reasoning-driven segmentation in remote sensing under a training-free setting to circumvent the supervision bottleneck, and introduce a dedicated benchmark to enable rigorous evaluation and diagnosis.

\begin{figure*}[t]
    \centering
    \setlength{\fboxsep}{0pt}
    \includegraphics[width=0.94\textwidth]{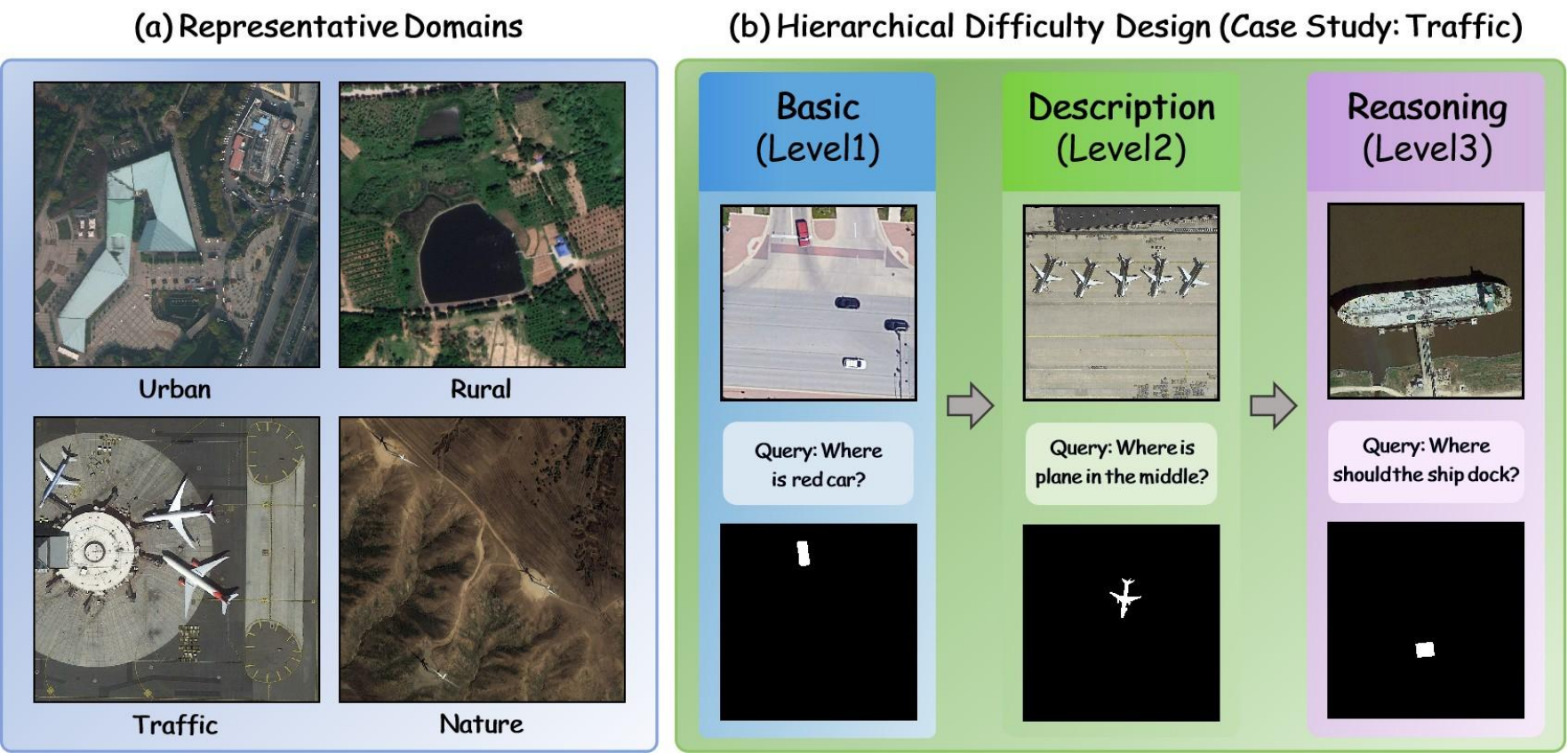}
    
    \caption{\textbf{Overview of GeoSeg-Bench.} 
    (a) \textbf{Representative Domains}: We showcase samples from the four domains defined in our scenario taxonomy: \textit{Urban}, \textit{Rural}, \textit{Traffic}, and \textit{Nature}. 
    (b) \textbf{Hierarchical Difficulty Design}: Using the \textit{Traffic} domain as a case study, we illustrate the progression across three levels: \textit{Basic} (Level 1), \textit{Description} (Level 2), and \textit{Reasoning} (Level 3), corresponding to increasing reasoning requirements.}
    \label{fig:bench_cases}
\end{figure*}

\section{Method}
\label{sec:method}

We present \textbf{GeoSeg}, a training-free framework for reasoning-driven segmentation in remote sensing imagery.
GeoSeg composes pretrained large models to translate open-ended instructions into precise pixel-level masks.
This section is organized following the pipeline in Figure~\ref{fig:pipeline}.
Sec.~\ref{sec:method_overview} defines the problem and framework overview;
Sec.~\ref{sec:method_grounding} details the \textbf{Reasoning-Driven Grounding} and \textbf{Bias-Aware Coordinate Refinement};
Sec.~\ref{sec:method_segmentation} elaborates on the \textbf{Dual-Route Segmentation} mechanism and our consensus-driven fusion strategy.

\subsection{Problem Definition and Framework Overview}

\label{sec:method_overview}

Let $I \in \mathbb{R}^{H \times W \times 3}$ be a remote sensing image and $q$ be an open-ended natural language query that may involve attributes, spatial relations, or implicit intent (e.g., \textit{``the residential buildings arranged in rows next to the park''}).
The goal is to predict a binary mask $M \in \{0,1\}^{H \times W}$, where $M_{u,v}=1$ indicates pixels belonging to the target region implied by $q$.

\noindent\textit{\textbf{Pipeline Overview.}}
Given $(I,q)$, GeoSeg proceeds in three sequential stages (see Figure~\ref{fig:pipeline}):
\begin{itemize}
    \item \textbf{Stage 1: Reasoning-Driven Grounding.} A multimodal large language model (MLLM) $\mathcal{L}$ analyzes the query $q$ to produce an initial candidate bounding box $b$ and a concise object prompt $p$.
    \item \textbf{Stage 2: Bias-Aware Coordinate Refinement.} To mitigate systematic grounding bias and localization uncertainty, we calibrate $b$ using a lightweight statistical correction, yielding a refined RoI.
    \item \textbf{Stage 3: Dual-Route Segmentation.} Within the RoI, we execute two parallel segmentation paths---a \textit{Point-Prompt Route} (via CLIP Surgery keypoints) and a \textit{Text-Prompt Route} (via SAM3 inference). The final mask is derived via a consensus-driven fusion mechanism.
\end{itemize}

GeoSeg is \emph{training-free} in the sense that we perform \textbf{no weight updates} to any component model:
all components ($\mathcal{L}=$ Qwen3-VL-32B~\cite{qwen3}, $\mathcal{C}=$ CLIP Surgery~\cite{clipsurgery}, $\mathcal{S}=$ SAM3~\cite{sam3})
are employed for zero-shot inference with their official pre-trained weights.
We only estimate \textbf{fixed geometric bias constants} $(\alpha,\beta)$ on a small \textbf{off-benchmark} held-out calibration set
(Sec.~\ref{sec:method_grounding}), without gradient-based learning and without using any GeoSeg-Bench samples.

\subsection{Reasoning-Driven Grounding and Bias-Aware Coordinate Refinement}

\label{sec:method_grounding}

\noindent\textit{\textbf{Reasoning-Driven Grounding.}}
The reasoning module $\mathcal{L}$ serves as the initial interpreter. We employ a composite prompt to guide $\mathcal{L}$ in decomposing the complex query $q$ into structured spatial and semantic outputs (see Appendix for full prompt details).
Given $(I,q)$, $\mathcal{L}$ generates a textual response from which we parse a grounding tuple:
\begin{equation}
    (b, p) = \mathrm{Parse}(\mathcal{L}(I, q)),
\end{equation}
where $b=[x_1,y_1,x_2,y_2]$ denotes the coarse bounding box in absolute pixel coordinates, and $p$ is a concise referential phrase extracted from $q$.
This step bridges the gap between high-level reasoning logic and pixel-level spatial localization.

\noindent\textit{\textbf{Bias-Aware Coordinate Refinement.}}
Directly utilizing the raw bounding box $b$ is suboptimal, as MLLMs pre-trained on natural images exhibit systematic coordinate misalignment when transferring to the overhead domain.
To quantify this, we constructed a \emph{held-out calibration set} by randomly sampling 1,000 high-quality images from the LoveDA, NWPU-VHR-10, and DIOR datasets, strictly distinct from GeoSeg-Bench.
As visualized in Figure~\ref{fig:bbox_bias}, the error distribution reveals a consistent bottom-right drift in predictions, which we attribute to the model's uncertainty in handling rotation-invariant overhead objects compared to gravity-aligned natural objects.
To rectify this, we apply an asymmetric statistical calibration.
We first clamp $b$ to image bounds to ensure validity, then expand it with bias-aware margins:
\begin{align}
\label{eq:expand}
w &= x_2-x_1,\quad h = y_2-y_1,\\
x_1' &= \mathrm{clip}(x_1-\alpha w,\,0,\,W),\quad
y_1' = \mathrm{clip}(y_1-\alpha h,\,0,\,H),\nonumber\\
x_2' &= \mathrm{clip}(x_2+\beta w,\,0,\,W),\quad
y_2' = \mathrm{clip}(y_2+\beta h,\,0,\,H),\nonumber
\end{align}
where $\alpha=0.2$ and $\beta=0.1$ are statistically derived to align with the offset distribution observed in our calibration analysis.
This yields a refined crop $I_{b'}$ that improves target coverage while limiting excessive background inclusion.

\begin{figure}[t]
    \centering
    \includegraphics[width=0.48\textwidth]{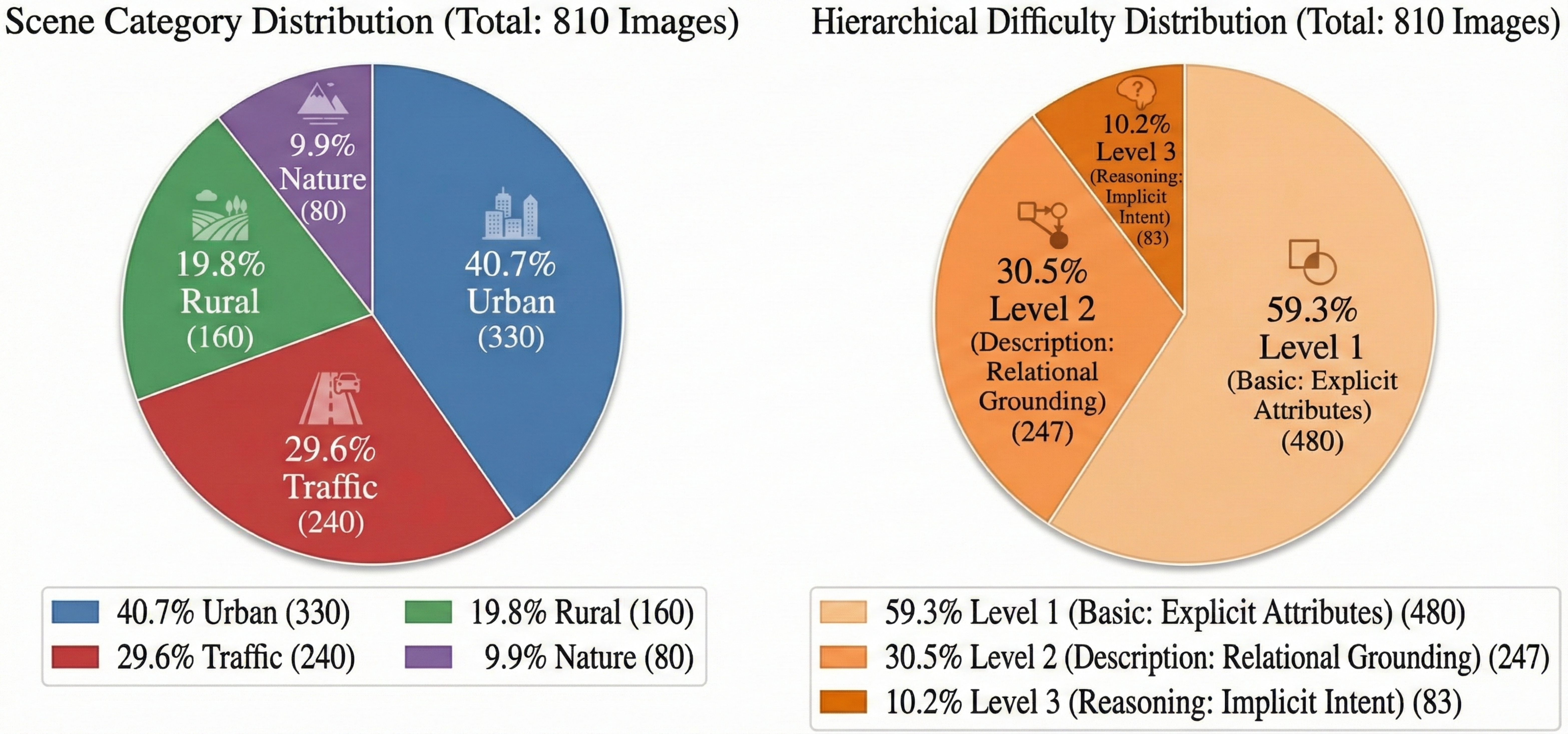}
    \vspace{-10pt}
    \caption{\textbf{Distribution statistics of our GeoSeg-Bench.} Left: Proportional breakdown of four scenario categories (Urban, Traffic, Rural, Nature). Right: Fixed ratio composition of three hierarchical difficulty levels (Basic, Description, Reasoning).}
    \label{fig:benchmark_rate}
\end{figure}

\subsection{Dual-Route Segmentation and Fusion}
\label{sec:method_segmentation}

To robustly segment the target within the refined crop $I_{b'}$, we introduce a Dual-Route mechanism.
This design leverages the complementarity between visual feature matching (Route A) and semantic text prompting (Route B).

\noindent\textit{\textbf{Route A: Point-Prompt Segmentation (Visual Keypoints).}}
This route mines visual cues to guide the segmenter. We employ a vision--language matcher $\mathcal{C}$ (specifically CLIP Surgery~\cite{clipsurgery}) to compute a similarity map $\Phi = \mathcal{C}(I_{b'}, p)$ between the crop and the prompt $p$.
We select CLIP Surgery over standard CLIP~\cite{CLIP} because the latter lacks fine-grained localization ability; in contrast, CLIP Surgery generates explainability maps that precisely highlight specific regions, thereby providing high-quality point prompts.
We extract these prompts by selecting at most $k$ local maxima from $\Phi$ via NMS:
\begin{equation}
    \mathcal{K} = \mathrm{TopK\text{-}NMS}(\Phi;\,k,\tau),
\end{equation}
where $\tau$ filters low-confidence responses and $k$ caps the number of points. Unless stated otherwise, we use a fixed setting of $k=5$ and $\tau=0.3$ for all experiments.
We then feed $\mathcal{K}$ to the segmenter $\mathcal{S}$ to obtain the point-prompted mask:
\begin{equation}
    \hat{M}^{\text{pt}}_{b'} = \mathcal{S}(I_{b'}, \mathcal{K}).
\end{equation}
If $\mathcal{K}=\emptyset$, we treat Route A as invalid in the fusion stage.
This route excels at focusing on salient object parts but depends on the quality of keypoint extraction.

\begin{figure*}[t]
\centering
\includegraphics[width=\textwidth]{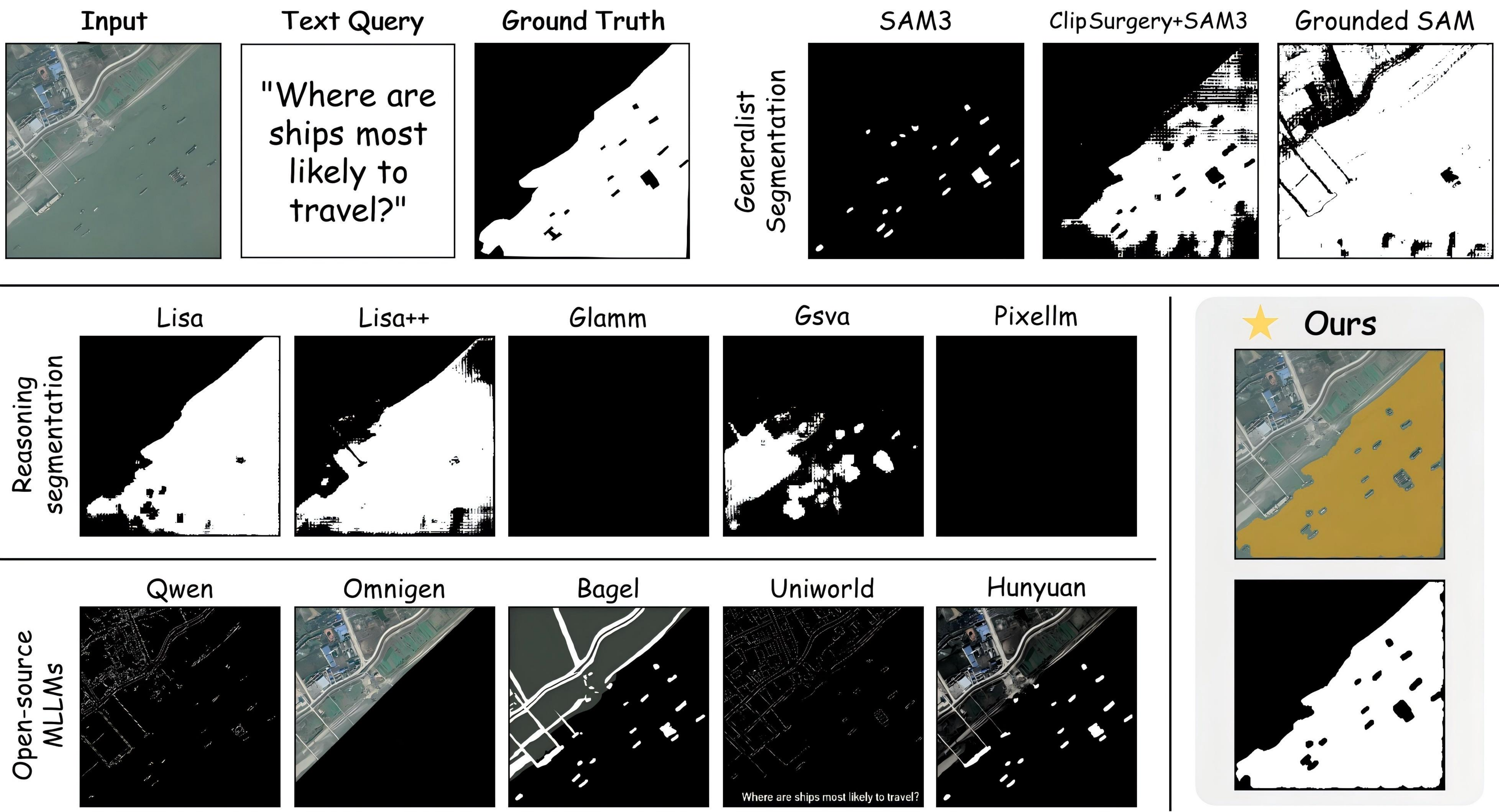}
\caption{\textbf{Qualitative comparison with multiple baselines.} This figure demonstrates the superiority of our approach over three major categories of baseline models: generalist segmentation, reasoning segmentation, and open-source MLLMs. As shown, most baseline methods struggle to comprehend the query intent, resulting in segmentation failures or excessive noise, whereas our method successfully generates accurate masks. More examples are in Appendix ~\ref{app:baseline}.}
\label{fig:compare}
\end{figure*}

\noindent\textit{\textbf{Route B: Text-Prompt Segmentation.}}
In parallel, this route utilizes the semantic capability of $\mathcal{S}$ directly. We feed the phrase $p$ as a text prompt:
\begin{equation}
    \hat{M}^{\text{txt}}_{b'} = \mathcal{S}(I_{b'}, p).
\end{equation}
This route captures global object context but may over-segment adjacent instances if the crop is loose.

\noindent\textit{\textbf{Consensus-Driven Fusion.}}
Both crop-level masks ($\hat{M}^{\text{pt}}_{b'}, \hat{M}^{\text{txt}}_{b'}$) are mapped back to the original image coordinates to obtain $\hat{M}^{\text{pt}}$ and $\hat{M}^{\text{txt}}$.
Concretely, we resize crop-level binary masks to the RoI resolution (nearest-neighbor) and paste them back to the original canvas using the RoI offsets.

To harmonize the precision of point prompts with the recall of text prompts, we adopt a consensus-driven fusion strategy.
Given the complex backgrounds in RS imagery, we deliberately prioritize the \emph{intersection} of the two routes. This strict consensus suppresses false positives arising from background clutter (common in Route B) or ambiguous keypoints (Route A). We apply the intersection only when \textbf{both} routes provide sufficiently reliable evidence; otherwise, we fall back to the \emph{single valid} route to avoid empty outputs while remaining conservative.

Let the validity indicator $\mathcal{V}(M)$ be defined as:
\begin{equation}
    \mathcal{V}(M) =  \mathbb{I}\!\left(\frac{A(M)}{A(b')} \ge \gamma\right),
\end{equation}
where $A(\cdot)$ denotes the pixel area and $A(b')$ is the RoI area. We set $\gamma=0.01$ as a small area-ratio threshold to filter degenerate masks while tolerating small objects.
 For the point-prompt route, we additionally set $\mathcal{V}(\hat{M}^{\text{pt}})=0$ when $\mathcal{K}=\emptyset$ (i.e., no valid keypoints are found).

The final prediction $\hat{M}$ is derived as:
\begin{equation}
\label{eq:fusion}
\hat{M}=
\begin{cases}
\hat{M}^{\text{pt}} \cap \hat{M}^{\text{txt}}, & \text{if } \mathcal{V}(\hat{M}^{\text{pt}}) \land \mathcal{V}(\hat{M}^{\text{txt}}),\\
\hat{M}^{\text{pt}}, & \text{if } \mathcal{V}(\hat{M}^{\text{pt}})\quad \text{(Fallback Route A)},\\
\hat{M}^{\text{txt}}, & \text{if } \mathcal{V}(\hat{M}^{\text{txt}})\quad \text{(Fallback Route B)},\\
\mathbf{0}, & \text{otherwise}.
\end{cases}
\end{equation}
This formulation ensures that we only output a mask when there is strong evidence, significantly enhancing robustness against distractors as verified in our ablation study (Table~\ref{tab:ablation}).

\section{GeoSeg-Bench: Benchmark Construction}
\label{sec:benchmark}

To enable rigorous evaluation of \emph{reasoning-driven segmentation} in remote sensing, we establish \textbf{GeoSeg-Bench}, a curated benchmark designed to pair standardized overhead imagery (predominantly $810{\times}810$ and $1024{\times}1024$) with open-ended natural language queries and pixel-accurate masks. Unlike conventional closed-set benchmarks relying on fixed labels, GeoSeg-Bench serves as a diagnostic testbed: it evaluates whether a model can interpret diverse instructions, ranging from explicit visual descriptions to implicit functional intent, and ground them into precise segmentation masks. Fig.~\ref{fig:bench_cases} illustrates the diversity of query styles and reasoning levels, while Fig.~\ref{fig:benchmark_rate} presents the comprehensive distribution statistics regarding scenarios and difficulty levels.

\subsection{Data Collection and Annotation Protocol}
\label{sec:bench_collect}

\textbf{\textit{Image Sources and Diversity.}}
To ensure the benchmark reflects real-world visual statistics, we aggregate imagery from diverse public datasets (including LoveDA~\cite{loveda}, Potsdam~\cite{potsdam}, NWPU-VHR-10~\cite{NWPU-VHR-10}, DIOR~\cite{DIOR}) and supplementary internet-sourced satellite/aerial images. This multi-source collection strategy covers a wide range of sensor types, varying Ground Sample Distances (GSD), and distinct scene layouts, effectively mitigating dataset bias.

\noindent\textbf{\textit{Manual Annotation and Quality Assurance.}}
For each image, we manually curate an instance triplet $(I,q,M)$. The queries $q$ are crafted to be open-ended and instruction-style, avoiding dataset-specific label taxonomies and encouraging natural descriptions.
Common nouns (e.g., \textit{lake}, \textit{hospital}) may appear, but queries are designed to require interpreting attributes, relations, or intent beyond a single class-name lookup. The ground-truth masks $M$ are annotated at the pixel level to enable fine-grained evaluation. To ensure high quality, all annotations underwent a rigorous verification process to eliminate ambiguities. The final benchmark comprises \textbf{810 images} with fully verified query-mask pairs.

\subsection{Scenario Taxonomy and Dataset Composition}
\label{sec:bench_scenario}

GeoSeg-Bench categorizes scenes into four representative domains (see Fig.~\ref{fig:benchmark_rate}, Left), reflecting common remote sensing applications and their unique challenges.
\textbf{Urban (330 images):} Dense built environments including commercial facilities, structured residential blocks, and industrial zones, characterized by high object density and complex shadows.
\textbf{Rural (160 images):} Agriculture-dominated scenes covering various farmland patterns, irrigation systems, and greenhouses, challenging models with texture homogeneity and seasonal variations.
\textbf{Traffic (240 images):} Transportation networks such as highways, bridges, intersections, airports, and parking lots, featuring extreme aspect ratios and connectivity reasoning.
\textbf{Nature (80 images):} Natural landscapes including water bodies, forests, bare land, and mountainous terrain, involving irregular boundaries and large-scale variation.

\subsection{Hierarchical Difficulty Design}
\label{sec:bench_level}

A core contribution of GeoSeg-Bench is its \textbf{hierarchical difficulty design}, which disentangles different capabilities required for reasoning segmentation. We stratify queries into three levels (Ratio: 60\% L1, 30\% L2, 10\% L3, as shown in Fig.~\ref{fig:benchmark_rate}, Right) to enable granular model diagnosis.

\textbf{Level 1 (Basic: Explicit Attributes).}
Level 1 queries focus on direct visual recognition based on explicit appearance cues, such as specific color, texture, and simple shape, sometimes paired with a common object noun for improved readability (e.g., \textit{``where is the blue lake?''}, \textit{``the circular irrigation pivot''}). Success here indicates a reliable foundational vision-language alignment capability.

\textbf{Level 2 (Description: Spatial \& Relational Grounding).}
Level 2 emphasizes spatial relations and layout-aware reasoning, requiring models to disambiguate specific targets within dense overhead scenes. Queries typically involve directional predicates like \textit{next to}, \textit{between}, or \textit{surrounded by} (e.g., \textit{``residential buildings arranged in rows next to the park''}). This tests the model's ability to understand geometric context and structural dependencies.

\textbf{Level 3 (Reasoning: Implicit Intent \& Causal Semantics).}
Level 3 represents the hardest tier, targeting implicit reasoning where the target is not explicitly named. These queries require functional reasoning (e.g., \textit{``Where can I seek medical help?''} $\to$ hospitals), state inference (e.g., \textit{``crops ready for harvest''} $\to$ yellow/mature fields), or causal/process reasoning (e.g., \textit{``Where does the river flow into?''} $\to$ river mouth). These queries stress-test the model's capacity to connect visual evidence with external world knowledge, a capability largely absent in traditional segmentation models.

\begin{figure*}[t]
    \centering
    \includegraphics[width=\textwidth]{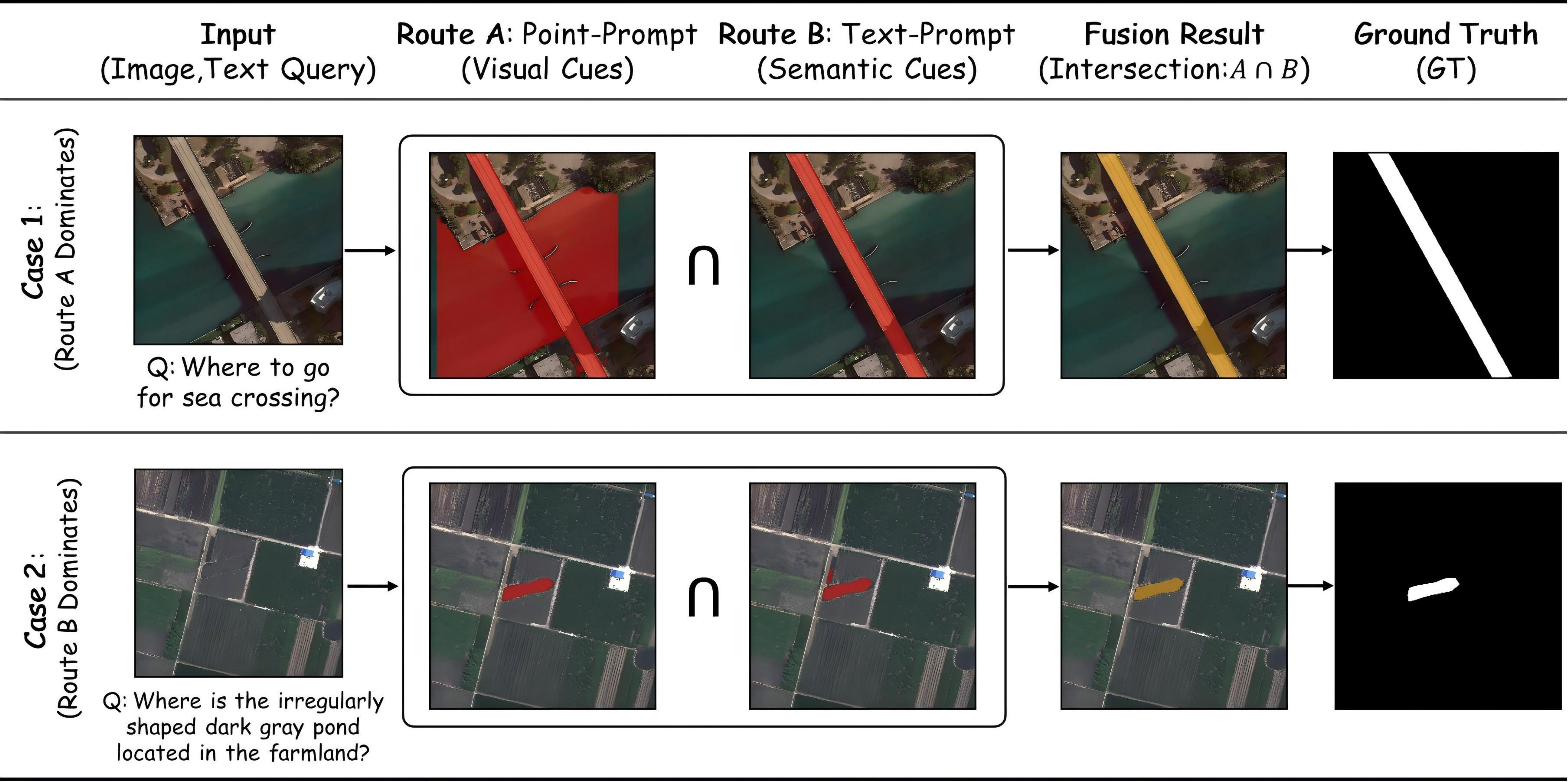}
    \caption{\textbf{Ablation study on component effectiveness.} 
    We validate the contribution of the Bias-Aware Coordinate Refinement (\textit{Box Refine}) and the Dual-Route strategy. 
    \textit{Route A} represents the Point-Prompt path (Visual Cues), and \textit{Route B} denotes the Text-Prompt path (Semantic Cues). 
    Removing any module significantly degrades performance, validating the necessity of our full pipeline. More examples in Appendix~\ref{app:ablation}.}

    \label{fig:ablation}
\end{figure*}

\begin{table*}[t]
\centering
\renewcommand{\arraystretch}{0.95} 
\caption{\textbf{Quantitative comparison on GeoSeg-Bench and the SegEarth-R2 training dataset.} 
We benchmark \textbf{GeoSeg} against representative methods from three categories: 
(1) \textit{Generalist segmentation}, 
(2) \textit{Reasoning segmentation}, and 
(3) \textit{Open-source MLLMs}. 
We report seven pixel-level metrics (\%) and inference speed (FPS). Best and second-best results are highlighted in \textbf{bold} and \underline{underlined}, respectively. Methods that fail to segment any target (IoU=0) are excluded from Accuracy and Specificity rankings. Inference speed is evaluated on GeoSeg-Bench with a single A100 GPU.}

\label{tab:main_results}
\resizebox{\textwidth}{!}{ 
\begin{tabular}{l *{8}{c} @{\hspace{1.5em}} *{7}{c}}
\toprule
\multirow{2}{*}{Method} & \multicolumn{8}{c}{GeoSeg-Bench ($\uparrow$)} & \multicolumn{7}{c}{SegEarth-R2 training dataset ($\uparrow$)} \\
\cmidrule(lr){2-9} \cmidrule(lr){10-16}
& IoU & Dice & Acc. & Prec. & Rec. & Spec. & BF & FPS & IoU & Dice & Acc. & Prec. & Rec. & Spec. & BF \\
\midrule

\multicolumn{16}{l}{\textbf{\textit{Generalist segmentation}}} \\
SAM3~\cite{sam3} 
& 19.9 & 26.8 & 89.3 & 21.7 & 58.6 & 91.8 & 12.9 & - 
& 10.5 & 14.3 & 77.0 & 11.2 & 47.3 & 78.0 & 10.9 \\
Grounded SAM~\cite{groundsam} 
& 18.6 & 25.1 & 64.2 & 20.0 & 73.1 & 63.9 & 10.9 & - 
& 11.3 & 15.1 & 61.9 & 13.1 & 55.2 & 61.8 & 11.1 \\
CLIP Surgery + SAM3~\cite{clipsurgery} 
& 24.7 & 31.4 & 73.9 & 26.8 & 69.8 & 74.2 & 10.4 & - 
& 14.5 & 19.0 & 69.0 & 15.5 & 55.7 & 69.3 & 10.1 \\
\midrule

\multicolumn{16}{l}{\textbf{\textit{Reasoning segmentation}}} \\
GLAMM~\cite{glamm} 
& 0.0 & 0.0 & 94.6 & 0.0 & 0.0 & 100.0 & 0.0 & - 
& 0.0 & 0.0 & 95.4 & 0.0 & 0.0 & 100.0 & 0.0 \\
GSVA~\cite{gsva} 
& 9.0 & 12.5 & 78.3 & 9.9 & 31.8 & 81.2 & 2.9 & - 
& 7.5 & 10.9 & 79.9 & 8.9 & 32.8 & 82.3 & 3.7 \\
LISA-7B~\cite{lisa} 
& \underline{39.5} & \underline{47.2} & 94.1 & \underline{42.7} & 60.9 & 95.4 & \underline{16.0} & 0.392
& \textbf{26.0} & \textbf{32.9} & 92.4 & \underline{28.4} & 55.2 & 93.4 & \textbf{16.9} \\
LISA++~\cite{lisa++} 
& 12.4 & 14.4 & \underline{95.2} & 14.0 & 16.8 & \textbf{99.2} & 5.1 & -
& 3.7 & 4.4 & \textbf{95.3} & 4.3 & 6.1 & \textbf{99.4} & 2.65 \\
PixelLM~\cite{pixellm} 
& 0.0 & 0.0 & 94.6 & 0.0 & 0.0 & 100.0 & 0.0 & 0.175
& 2.4 & 3.1 & 95.4 & 3.5 & 4.1 & 99.8 & 1.3 \\
\midrule

\multicolumn{16}{l}{\textbf{\textit{Open-source MLLMs}}}\\
Qwen-Edit~\cite{qwenimage} 
& 8.7 & 13.9 & 49.6 & 10.0 & 68.7 & 48.4 & 2.4 & 0.011
& 5.8 & 9.7 & 57.3 & 7.2 & 58.1 & 57.2 & 3.9 \\
OmniGen2~\cite{omnigen2} 
& 6.8 & 10.8 & 43.3 & 7.2 & 67.4 & 42.1 & 2.7 & 0.011
& 5.8 & 9.3 & 34.8 & 6.5 & 74.2 & 33.4 & 3.6 \\
BAGEL~\cite{bagel} 
& 10.7 & 16.2 & 57.2 & 11.3 & 74.7 & 56.3 & 3.5 & 0.001
& 8.8 & 13.4 & 43.8 & 9.2 & \underline{85.4} & 42.0 & 5.4 \\
UniWorld~\cite{uniworld} 
& 8.8 & 14.0 & 44.0 & 9.2 & \underline{78.3} & 41.6 & 2.4 & 0.015
& 6.4 & 10.3 & 47.2 & 7.1 & 70.0 & 45.7 & 3.8 \\
Hunyuan~\cite{hunyuanimage} 
& 5.6 & 9.7 & 10.8 & 5.6 & \textbf{99.2} & 5.7 & 1.0 & 0.002 
& 4.3 & 7.6 & 12.0 & 4.3 & \textbf{98.1} & 8.2 & 1.7 \\
\midrule

\multicolumn{16}{l}{\textbf{\textit{Ours}}} \\
\textbf{GeoSeg}
& \textbf{56.4} & \textbf{64.2} & \textbf{96.8} & \textbf{69.0} & 65.7 & \underline{98.3} & \textbf{26.6} & 0.221 
& \underline{17.4} & \underline{22.0} & \underline{94.7} & \textbf{29.2} & 25.2 & \underline{97.0} & \underline{13.2} \\
\bottomrule
\end{tabular}
}
\end{table*}

\section{Experiments}
\label{sec:experiments}

\subsection{Experimental Settings}
\label{sec:exp_settings}

\noindent\textbf{\textit{Datasets and Zero-Shot Evaluation Protocol.}}
We evaluate our method using two data sources: our newly established \textbf{GeoSeg-Bench} (Sec.~\ref{sec:benchmark}) and \textbf{SegEarth-R2}~\cite{segearth}. Although proposed as a large-scale training corpus, SegEarth-R2's 10,000 reasoning segmentation cases in remote sensing imagery are repurposed as a benchmark (details in Appendix ~\ref{app:benchmark_dataset}). Crucially, to validate the \emph{training-free} advantage and generalization of our approach, we enforce a strict zero-shot testing protocol. In stark contrast to frameworks like LISA and PixelLM that rely on large-scale reasoning segmentation datasets for training, our method has never been trained or fine-tuned on such data. Furthermore, none of the evaluated models have seen GeoSeg-Bench or SegEarth-R2 during training. All methods are evaluated in pure zero-shot inference mode, preventing data leakage and genuinely reflecting reasoning capabilities in unseen environments.

\noindent\textbf{\textit{Evaluation Metrics and MLLM-Judge.}}
For quantitative spatial analysis, we report seven standard pixel-level metrics: Intersection over Union (IoU), Dice coefficient, Accuracy, Precision, Recall, Specificity, and Boundary F-score (BF). Furthermore, recognizing that pure pixel overlap can penalize functionally correct but morphologically varied predictions, we employ an \textbf{MLLM-as-a-judge} protocol (powered by Qwen3-VL-8B/32B). This protocol assesses the semantic alignment of the segmentation outputs on a 1--5 Likert scale across three criteria: \textit{Faithfulness} (instruction adherence), \textit{Localization} (boundary precision), and \textit{Robustness} (distractor avoidance). The exact prompt templates and detailed grading guidelines designed for the MLLM judge are provided in Appendix~\ref{sec:mllm_judge}.

\noindent\textbf{\textit{Baselines.}}
We benchmark our approach against \textbf{13} representative state-of-the-art baselines. These span three major paradigms: (i) generalist segmentation models (e.g., SAM3, Grounded SAM), (ii) reasoning segmentation frameworks (e.g., LISA, PixelLM), and (iii) visual-language foundation models/MLLMs. To ensure fairness under our training-free protocol, all 13 baselines are deployed using their official pre-trained weights without any target-domain adaptation, and are queried using standardized prompts consistent with our method.

\noindent\textbf{\textit{Ablation Study Design.}}
To dissect the efficacy of our proposed architecture, we establish the full GeoSeg framework as the anchor and systematically ablate its core modules (Table~\ref{tab:ablation}):
(i) \textbf{w/o Box Refinement}: We bypass the asymmetric coordinate correction module, directly feeding raw VLM bounding boxes to the segmenter. This quantifies the impact of the systematic spatial shift inherent in VLM grounding.
(ii) \textbf{Single-Route Variants}: We isolate the parallel branches of our Dual-Route mechanism. We evaluate the performance when relying exclusively on the Point-Prompt path (\textit{Route A}, disabling Text-Prompts) and vice versa (\textit{Route B}, disabling Point-Prompts). This evaluates the necessity of synergizing coarse semantic guidance with fine-grained visual cues.

\begin{table*}[t]
\centering
\scriptsize 
\renewcommand{\arraystretch}{1.1} 
\caption{\textbf{Model-based and human evaluation on GeoSeg-Bench and the SegEarth-R2 training dataset.}
We assess reasoning-driven segmentation quality using MLLM judges (Qwen3-VL-8B/32B) and human evaluators (User).
Predictions are rated on a 1--5 scale and averaged.
Metrics: \textit{Faithfulness} (F.; compliance with constraints), \textit{Localization} (L.; correctness of mask boundaries), and \textit{Robustness} (R.; avoidance of distractors). Best and second-best are \textbf{bold} and \underline{underlined}, respectively.}
\label{tab:mllm_results}

\resizebox{\textwidth}{!}{
\begin{tabular}{l *{3}{c} @{\hspace{1em}} *{3}{c} @{\hspace{1em}} *{3}{c} @{\hspace{1.5em}} *{3}{c} @{\hspace{1em}} *{3}{c} @{\hspace{1em}} *{3}{c}}
\toprule
\multirow{3}{*}{Method} & \multicolumn{9}{c}{GeoSeg-Bench ($\uparrow$)} & \multicolumn{9}{c}{SegEarth-R2 training dataset ($\uparrow$)} \\
\cmidrule(lr){2-10} \cmidrule(lr){11-19}
& \multicolumn{3}{c}{Qwen-8B} & \multicolumn{3}{c}{Qwen-32B} & \multicolumn{3}{c}{User} 
& \multicolumn{3}{c}{Qwen-8B} & \multicolumn{3}{c}{Qwen-32B} & \multicolumn{3}{c}{User} \\
\cmidrule(lr){2-4} \cmidrule(lr){5-7} \cmidrule(lr){8-10} \cmidrule(lr){11-13} \cmidrule(lr){14-16} \cmidrule(lr){17-19}
& F. & L. & R. & F. & L. & R. & F. & L. & R. & F. & L. & R. & F. & L. & R. & F. & L. & R. \\
\midrule

\multicolumn{19}{l}{\textbf{\textit{Generalist segmentation}}} \\
SAM3~\cite{sam3} 
& 1.78 & 1.72 & 1.81 & 2.03 & 1.74 & 2.00 & 2.75 & 2.30 & 2.68 
& 1.39 & 1.36 & 1.40 & 1.50 & 1.37 & 1.46 & 2.00 & 1.87 & 1.99 \\
Grounded SAM~\cite{groundsam} 
& 1.71 & 1.65 & 1.75 & 1.92 & 1.64 & 1.86 & 2.63 & 2.17 & 2.50 
& 1.42 & 1.39 & 1.44 & 1.55 & 1.41 & 1.50 & 2.05 & 1.88 & 2.03 \\
CLIP Surgery~\cite{clipsurgery} 
& 1.98 & 1.91 & 2.02 & 2.15 & 1.88 & 2.05 & 2.86 & 2.51 & 2.78 
& 1.57 & 1.52 & 1.61 & 1.74 & 1.54 & 1.66 & 2.35 & 2.05 & 2.25 \\
\midrule

\multicolumn{19}{l}{\textbf{\textit{Reasoning segmentation}}} \\
GLAMM~\cite{glamm} 
& 1.00 & 1.00 & 1.00 & 1.00 & 1.00 & 1.00 & 1.38 & 1.30 & 1.38 
& 1.00 & 1.00 & 1.00 & 1.01 & 1.01 & 1.01 & 1.38 & 1.35 & 1.33 \\
GSVA~\cite{gsva} 
& 1.32 & 1.30 & 1.35 & 1.42 & 1.30 & 1.39 & 1.96 & 1.74 & 1.84 
& 1.25 & 1.23 & 1.28 & 1.37 & 1.25 & 1.33 & 1.81 & 1.72 & 1.81 \\
LISA-7B~\cite{lisa} 
& \underline{2.77} & \underline{2.65} & \underline{2.83} & \underline{2.97} & \underline{2.51} & \underline{2.70} & \underline{3.85} & \underline{3.45} & \underline{3.65} 
& \textbf{2.12} & \textbf{2.02} & \textbf{2.18} & \textbf{2.35} & \textbf{2.00} & \textbf{2.21} & \textbf{3.10} & \underline{2.45} & \textbf{2.95} \\
LISA++~\cite{lisa++} 
& 1.57 & 1.54 & 1.58 & 1.62 & 1.49 & 1.53 & 2.22 & 2.03 & 2.07 
& 1.16 & 1.15 & 1.16 & 1.18 & 1.14 & 1.16 & 1.64 & 1.53 & 1.57 \\
PixelLM~\cite{pixellm} 
& 1.04 & 1.04 & 1.04 & 1.05 & 1.03 & 1.04 & 1.45 & 1.40 & 1.44 
& 1.10 & 1.09 & 1.11 & 1.13 & 1.09 & 1.11 & 1.53 & 1.49 & 1.45 \\
\midrule

\multicolumn{19}{l}{\textbf{\textit{Open-source MLLMs}}}\\
Qwen-Edit~\cite{qwenimage} 
& 1.21 & 1.20 & 1.22 & 1.32 & 1.21 & 1.26 & 1.75 & 1.61 & 1.66 
& 1.12 & 1.11 & 1.13 & 1.13 & 1.11 & 1.14 & 1.50 & 1.46 & 1.52 \\
OmniGen2~\cite{omnigen2} 
& 1.15 & 1.14 & 1.17 & 1.21 & 1.15 & 1.22 & 1.65 & 1.54 & 1.63 
& 1.14 & 1.13 & 1.16 & 1.22 & 1.14 & 1.19 & 1.62 & 1.52 & 1.65 \\
BAGEL~\cite{bagel} 
& 1.31 & 1.29 & 1.34 & 1.43 & 1.31 & 1.42 & 2.20 & 1.95 & 2.15 
& 1.26 & 1.23 & 1.28 & 1.40 & 1.25 & 1.33 & 2.10 & 1.85 & 2.00 \\
UniWorld~\cite{uniworld} 
& 1.22 & 1.21 & 1.24 & 1.35 & 1.23 & 1.32 & 1.84 & 1.67 & 1.75 
& 1.15 & 1.14 & 1.17 & 1.26 & 1.16 & 1.21 & 1.72 & 1.53 & 1.62 \\
Hunyuan~\cite{hunyuanimage} 
& 1.06 & 1.06 & 1.06 & 1.10 & 1.07 & 1.11 & 1.53 & 1.46 & 1.50 
& 1.04 & 1.04 & 1.05 & 1.10 & 1.05 & 1.08 & 1.50 & 1.45 & 1.49 \\
\midrule

\multicolumn{19}{l}{\textbf{\textit{Ours}}} \\
\textbf{GeoSeg}
& \textbf{3.64} & \textbf{3.48} & \textbf{3.66} & \textbf{3.72} & \textbf{3.28} & \textbf{3.28} & \textbf{4.35} & \textbf{4.12} & \textbf{4.20} 
& \underline{1.78} & \underline{1.71} & \underline{1.81} & \underline{1.94} & \underline{1.69} & \underline{1.82} & \underline{2.65} & \textbf{2.80} & \underline{2.50} \\
\bottomrule
\end{tabular}
}

\end{table*}

\subsection{Main Results}
\label{sec:main_results}

\noindent\textbf{\textit{Pixel-Level Quantitative Evaluation.}}
Table~\ref{tab:main_results} summarizes pixel-level performance and inference speed. %
On our highly challenging GeoSeg-Bench, GeoSeg demonstrates overwhelming superiority, achieving the best performance across almost all metrics, including 56.4\% IoU and 64.2\% Dice. %
Crucially, despite being completely training-free, GeoSeg significantly surpasses the strongest reasoning baseline, LISA-7B (39.5\% IoU)---which relies on extensive domain-specific fine-tuning---and drastically outperforms generalist models like CLIP Surgery + SAM3 (24.7\% IoU). %
On the SegEarth-R2 training set, GeoSeg maintains highly competitive zero-shot transferability. %
While the fully supervised LISA-7B achieves the highest IoU here, GeoSeg secures the highest Precision (29.2\%), demonstrating its predictions are notably more accurate with fewer false positives. %
Furthermore, beyond segmentation accuracy, GeoSeg maintains high efficiency. It significantly outperforms all open-source MLLMs in terms of inference speed and is only slightly slower than a few reasoning segmentation methods.
Overall, GeoSeg proves to be a highly reliable and efficient framework for grounding complex queries into precise masks under a strict zero-shot protocol. %

\noindent\textbf{\textit{Semantic Alignment via MLLM/VLM-Judge.}}
Since traditional pixel metrics can penalize functionally correct but morphologically distinct predictions, we leverage MLLM judges (Table~\ref{tab:mllm_results}). On GeoSeg-Bench, \textbf{GeoSeg ranks \#1} across all methods. Evaluated by Qwen3-VL-8B, GeoSeg attains state-of-the-art scores in Faithfulness (\textbf{3.64}), Localization (\textbf{3.48}), and Robustness (\textbf{3.66}). We note that on the SegEarth-R2 training set, LISA-7B achieves higher judge scores. This gap is fully expected, as LISA explicitly trains on large-scale reasoning segmentation datasets, whereas GeoSeg operates in a strictly training-free manner. Despite this, our method vastly outperforms traditional MLLMs (e.g., Qwen-Edit, BAGEL) that score below 1.50, confirming GeoSeg effectively bridges high-level reasoning intent and spatial grounding without domain-specific training data.

\noindent\textbf{\textit{User Study.}}
To ensure our model's outputs align with human expectations, we conducted a user study involving 50 participants who blindly rated 40 randomly sampled image-query pairs (detailed testing protocols in Appendix~\ref{uers_study}). As shown in Table~\ref{tab:mllm_results}, GeoSeg overwhelmingly dominates the human preference rankings, achieving an exceptional Faithfulness score of \textbf{4.35}, Localization of \textbf{4.12}, and Robustness of \textbf{4.20}. Human evaluators noted our method's ability to accurately resolve ambiguous queries while strictly ignoring same-class distractors—a capability directly attributable to our Dual-Route design.

\begin{table*}[t]
\centering
\renewcommand{\arraystretch}{1}

\caption{\textbf{Ablation study of component effectiveness on GeoSeg-Bench and the SegEarth-R2 training dataset.}
We evaluate the contribution of the \textit{Box Refinement} module and the two parallel paths in our \textit{Dual-Route} mechanism.
\textit{Route A} denotes the \textbf{Point-Prompt} path, and \textit{Route B} denotes the \textbf{Text-Prompt} path.
Removing any component degrades overall performance. Best and second-best are \textbf{bold} and \underline{underlined}, respectively.}
\label{tab:ablation}
\resizebox{\textwidth}{!}{
\begin{tabular}{ccc @{\hspace{1.5em}} ccccccc @{\hspace{1.5em}} ccccccc}
\toprule
\multicolumn{3}{c}{Components} & \multicolumn{7}{c}{GeoSeg-Bench ($\uparrow$)} & \multicolumn{7}{c}{SegEarth-R2 training dataset ($\uparrow$)} \\
\cmidrule(lr){1-3} \cmidrule(lr){4-10} \cmidrule(lr){11-17}
Box Refine & Route A & Route B & IoU & Dice & Acc. & Prec. & Rec. & Spec. & BF & IoU & Dice & Acc. & Prec. & Rec. & Spec. & BF \\
\midrule

\xmark & \cmark & \cmark 
& 51.1 & 59.9 & \underline{96.5} & \underline{62.7} & 64.3 & 97.8 & 21.7 
& 13.4 & 17.7 & 93.9 & 22.9 & 21.1 & 96.2 & 9.7 \\

\cmark & \xmark & \cmark 
& \underline{52.9} & \underline{61.4} & 95.8 & 59.6 & \textbf{71.5} & 96.6 & \underline{24.3} 
& \underline{16.0} & \underline{20.6} & 93.1 & \underline{23.6} & \textbf{27.9} & 95.0 & \underline{11.6} \\

\cmark & \cmark & \xmark 
& 43.2 & 49.0 & \underline{96.5} & 52.4 & 48.9 & \textbf{99.0} & 21.2 
& 15.2 & 19.1 & \textbf{95.1} & 22.7 & 22.6 & \textbf{97.6} & 11.4 \\

\midrule
\cmark & \cmark & \cmark 
& \textbf{56.4} & \textbf{64.2} & \textbf{96.8} & \textbf{69.0} & \underline{65.7} & \underline{98.3} & \textbf{26.6} 
& \textbf{17.4} & \textbf{22.0} & \underline{94.7} & \textbf{29.2} & \underline{25.2} & \underline{97.0} & \textbf{13.2} \\
\bottomrule
\end{tabular}
}
\end{table*}

\subsection{Ablation Study}
\label{sec:ablation}

We isolate the contributions of core components on GeoSeg-Bench and the SegEarth-R2 dataset (Table~\ref{tab:ablation}, Fig.~\ref{fig:ablation}). 
Results confirm our complete GeoSeg pipeline is optimal; removing any single module causes severe performance drops across both benchmarks.
Specifically, disabling Box Refinement amplifies the raw VLM's localization bias, dropping the IoU on GeoSeg-Bench from 56.4\% to \textbf{51.1\%}. 
More critically, removing \textbf{Route B} (Text-Prompt path) triggers a substantial performance collapse (IoU plummets to \textbf{43.2\%}). Lacking explicit semantic descriptors, the model suffers severe \emph{background leakage}, failing to disambiguate targets from surroundings.
Conversely, removing \textbf{Route A} (Point-Prompt path) yields \textbf{52.9\%} IoU. Without fine-grained visual anchors, the model tends to \emph{over-segment same-class distractors} and merge adjacent distinct objects, which also leads to a noticeable degradation in boundary quality (BF drops to \textbf{24.3\%}). 
Collectively, these consistent ablations prove that precise coordinate correction and the synergy between visual (Route A) and semantic (Route B) cues are indispensable for reasoning-driven remote sensing segmentation.

\section{Conclusion}
\label{sec:conclusion}

We presented GeoSeg, a training-free framework for reasoning-driven remote sensing segmentation. 
To enable reliable zero-shot evaluation, we introduced GeoSeg-Bench, covering diverse image-query pairs. 
GeoSeg bridges VLM understanding and precise mask prediction via: (i) Box Refinement to correct systematic grounding bias, and (ii) a Dual-Route mechanism combining point-prompts (fine-grained localization) with text-prompts (semantic disambiguation). 
GeoSeg achieves state-of-the-art pixel-level performance and ranks \#1 in MLLM-as-a-judge and user studies, demonstrating superior instruction faithfulness, localization, and robustness. 
Ablations corroborate the necessity and complementarity of refined grounding and dual-route synergy.

GeoSeg inherits limitations from its underlying models, including occasional grounding failures, sensitivity to long-tail prompts, reliance on static refinement margins, and added inference cost. 
Future work will explore adaptive scale-aware calibration, uncertainty-aware refinement, and interactive correction loops. 
We also plan extensions to instance/panoptic segmentation and multi-temporal imagery, further improving scalability and real-world usability. GeoSeg establishes a new paradigm for resource-efficient remote sensing analysis, demonstrating that high-level reasoning does not inherently require high-cost supervision.

\newpage

\bibliography{example_paper}
\bibliographystyle{icml2025}

\newpage
\appendix
\onecolumn

\section{MLLM-as-a-Judge Prompt Template and Grading Guidelines}
\label{sec:mllm_judge}

To complement standard pixel-level metrics and evaluate the semantic alignment of our segmentation outputs, we employ an MLLM-as-a-judge protocol (powered by Qwen3-VL). Standard overlap metrics can sometimes penalize predictions that are functionally correct but morphologically varied. To address this, we cast the MLLM as an expert in remote sensing image analysis. 

The model is provided with a composite image containing the original background, the ground truth mask (rendered in green), the predicted mask (rendered in red), and the overlapping regions (rendered in yellow). It is then instructed to evaluate the segmentation quality across four distinct dimensions—Faithfulness, Localization, Robustness, and Overlap—using a 1-to-5 Likert scale. 

The exact prompt template and detailed scoring rubrics fed to the model are structured as follows:

\vspace{6pt}
\begin{promptbox}[System Prompt \& Scoring Rubric]
\small 
\textbf{You are an expert in remote sensing image analysis. Evaluate the segmentation quality of a [Class Name] in this remote sensing image.}

\vspace{4pt}
\noindent \textbf{The image shows:}
\begin{itemize}
    \setlength\itemsep{0em}
    \item Original image (background)
    \item Green overlay: Ground truth mask (correct segmentation)
    \item Red overlay: Predicted mask (model's segmentation)
    \item Yellow areas: Overlapping regions (where both masks agree)
\end{itemize}

\noindent \textbf{Image dimensions:} [Width] x [Height] pixels \\
\textbf{Class:} [Class Name]

\vspace{4pt}
\noindent Evaluate the segmentation from FOUR aspects and provide integer scores from 1 to 5 for each metric. Use the following scoring standards:

\vspace{4pt}
\noindent \textbf{1. Faithfulness:} Does the predicted mask correctly identify the [Class Name]?
\begin{itemize}
    \setlength\itemsep{0em}
    \item \textbf{5 (Excellent):} Perfectly correct identification, no confusion with other classes.
    \item \textbf{4 (Good):} Mostly correct, minor confusion with similar classes.
    \item \textbf{3 (Fair):} Generally correct but some confusion with related classes.
    \item \textbf{2 (Poor):} Significant confusion, partially wrong class identification.
    \item \textbf{1 (Very Poor):} Completely wrong class, major misidentification.
\end{itemize}

\noindent \textbf{2. Localization:} Does the predicted mask precisely follow the complex edges?
\begin{itemize}
    \setlength\itemsep{0em}
    \item \textbf{5 (Excellent):} Boundaries perfectly match ground truth, no rounded corners or overflow.
    \item \textbf{4 (Good):} Boundaries mostly accurate, minor deviations at complex edges.
    \item \textbf{3 (Fair):} Generally follows boundaries but noticeable deviations or slight overflow.
    \item \textbf{2 (Poor):} Significant boundary misalignment, obvious rounded corners or overflow.
    \item \textbf{1 (Very Poor):} Severe boundary errors, completely misaligned edges.
\end{itemize}

\noindent \textbf{3. Robustness:} Can the segmentation resist interference from clouds, shadows, seasonal changes, or similar textures?
\begin{itemize}
    \setlength\itemsep{0em}
    \item \textbf{5 (Excellent):} Highly robust, unaffected by environmental variations or similar textures.
    \item \textbf{4 (Good):} Mostly robust, minor sensitivity to environmental factors.
    \item \textbf{3 (Fair):} Moderate robustness, some sensitivity to clouds/shadows/similar textures.
    \item \textbf{2 (Poor):} Low robustness, easily confused by environmental variations.
    \item \textbf{1 (Very Poor):} Very fragile, severely affected by clouds, shadows, or similar textures.
\end{itemize}

\noindent \textbf{4. Overlap:} Pixel-level IoU (Intersection over Union) between predicted and ground truth masks.
\begin{itemize}
    \setlength\itemsep{0em}
    \item \textbf{5 (Excellent):} IoU $\geq$ 0.8, excellent pixel-level overlap.
    \item \textbf{4 (Good):} IoU 0.6--0.8, good overlap with minor differences.
    \item \textbf{3 (Fair):} IoU 0.4--0.6, moderate overlap, noticeable differences.
    \item \textbf{2 (Poor):} IoU 0.2--0.4, poor overlap, significant differences.
    \item \textbf{1 (Very Poor):} IoU $<$ 0.2, very poor overlap, minimal agreement.
\end{itemize}

\vspace{4pt}
\noindent \textit{(Note: The actual calculated IoU is provided to the model as a reference).}

\vspace{4pt}
\noindent Respond with ONLY a valid JSON object in this format (use integer scores 1-5): \\
\texttt{\{"faithfulness": <1-5>, "localization": <1-5>, "robustness": <1-5>, "overlap": <1-5>\}}
\end{promptbox}

\section{Additional Qualitative Comparisons with Baselines}
\label{app:baseline}

In this section, we provide five additional qualitative examples to further demonstrate the robustness and superiority of our proposed approach compared to existing baselines. Consistent with the evaluation in the main text, the comparisons span three major categories: generalist segmentation, reasoning segmentation, and open-source MLLMs. 

To comprehensively evaluate the models, these five selected cases encompass varying degrees of query complexity, visual ambiguity, and challenging object scales. As illustrated in Figure~\ref{fig:more_baselines}, the baseline models frequently fail to accurately ground the target objects. They tend to exhibit severe over-segmentation or miss the target entirely due to a limited comprehension of complex, implicit instructions. In contrast, our approach consistently demonstrates a deep understanding of the user's intent, effectively bridging the gap between complex textual queries and visual contexts to produce highly precise segmentation masks across all five challenging scenarios.

\begin{figure}[htbp]
    \centering
    \begin{subfigure}{\textwidth}
        \includegraphics[width=0.98\textwidth]{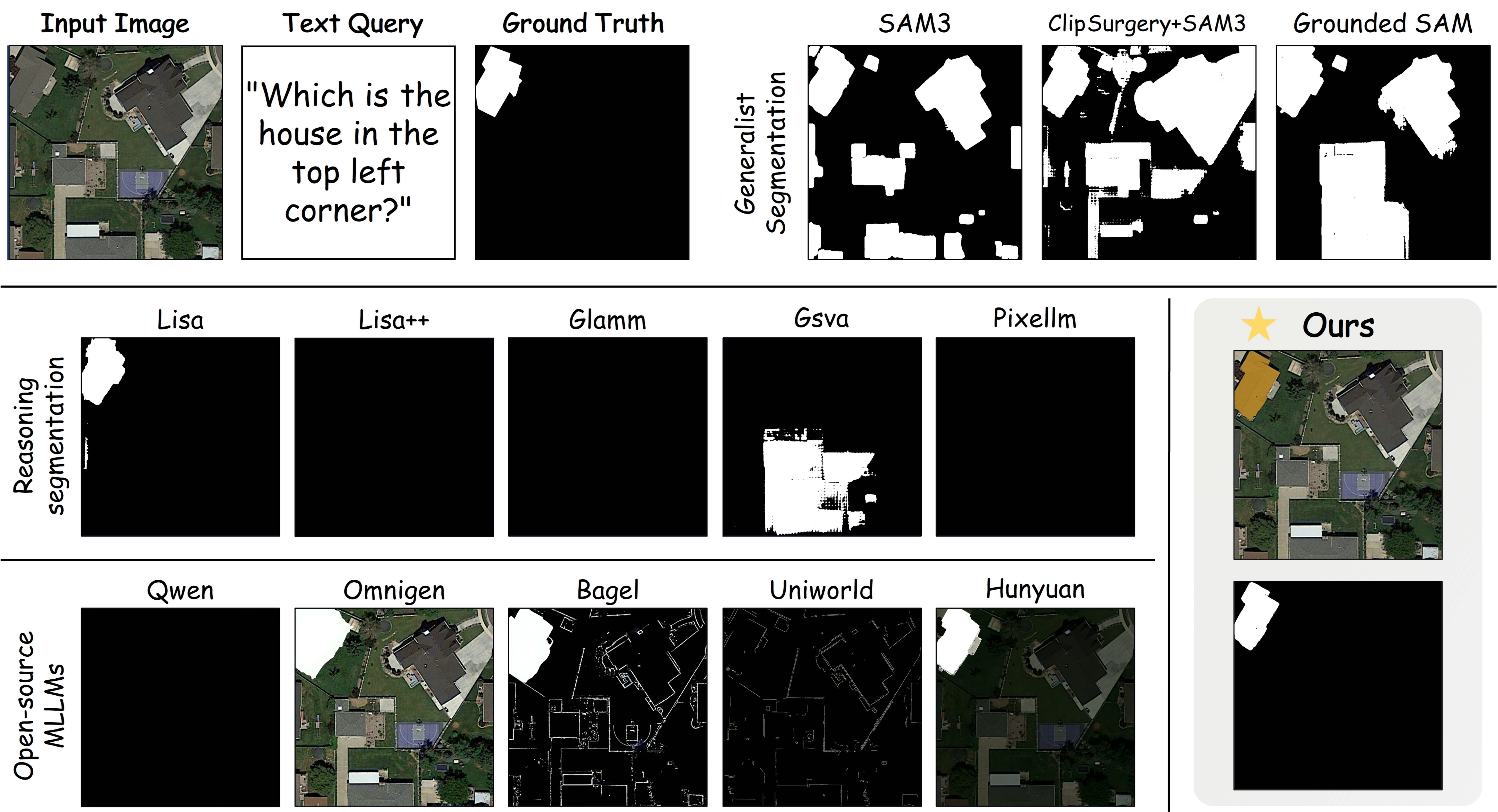}
        \caption{Case 1: Dealing with visual ambiguity.}
    \end{subfigure}
    \vspace{14pt} 
    
    \begin{subfigure}{0.98\textwidth}
        \includegraphics[width=\textwidth]{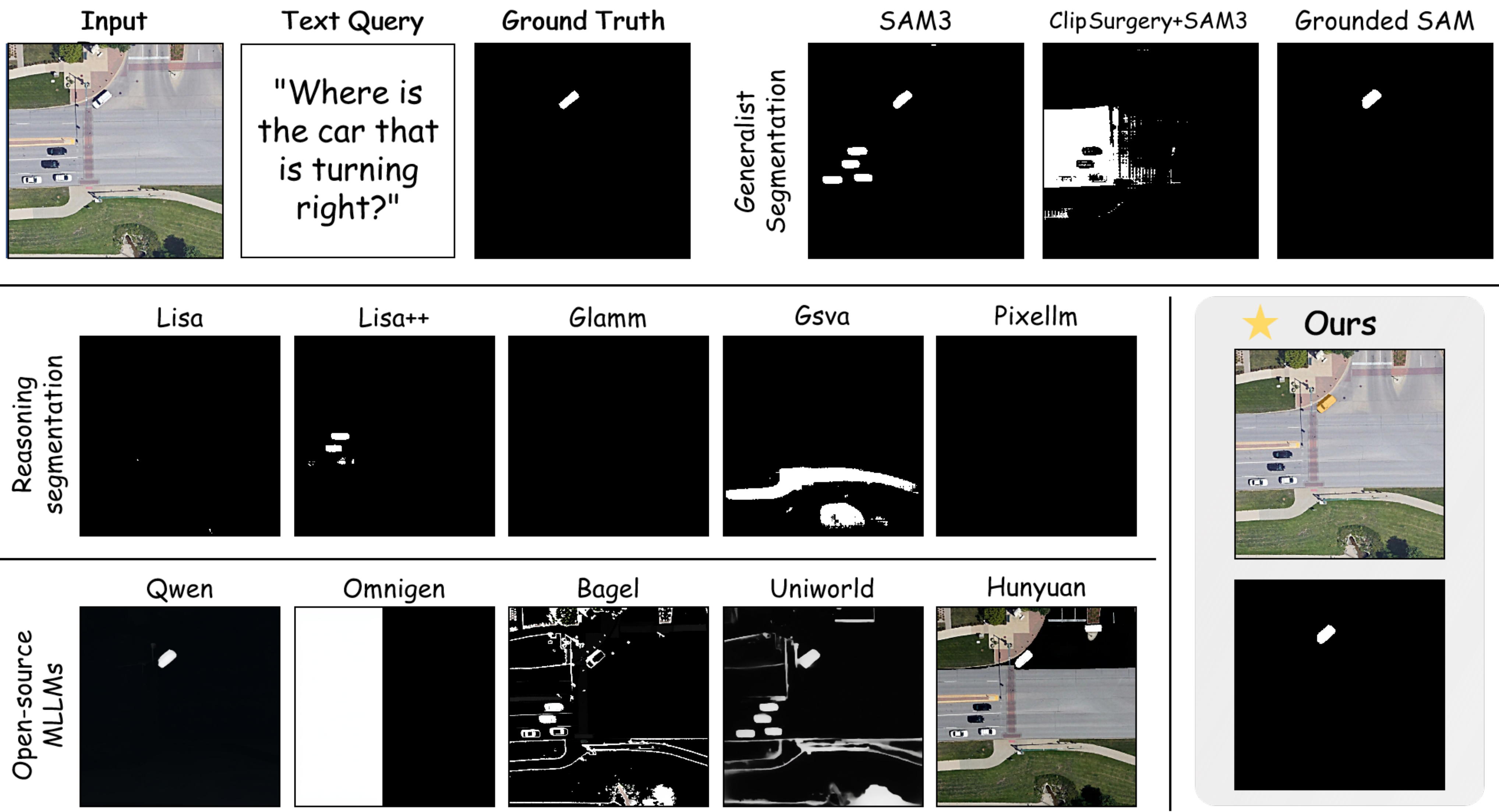}
        \caption{Case 2: Complex spatial reasoning.}
    \end{subfigure}
    \vspace{14pt}

    \caption{\textbf{Extended qualitative comparison with multiple baselines.} We present five additional challenging cases. Compared to generalist segmentation, reasoning segmentation, and open-source MLLMs, our method consistently interprets complex queries accurately and outputs precise masks. (Figure continued on next page.)}
    \label{fig:more_baselines}
\end{figure}

\begin{figure}[htbp]
    \ContinuedFloat 
    \centering
    \begin{subfigure}{0.98\textwidth}
        \includegraphics[width=\textwidth]{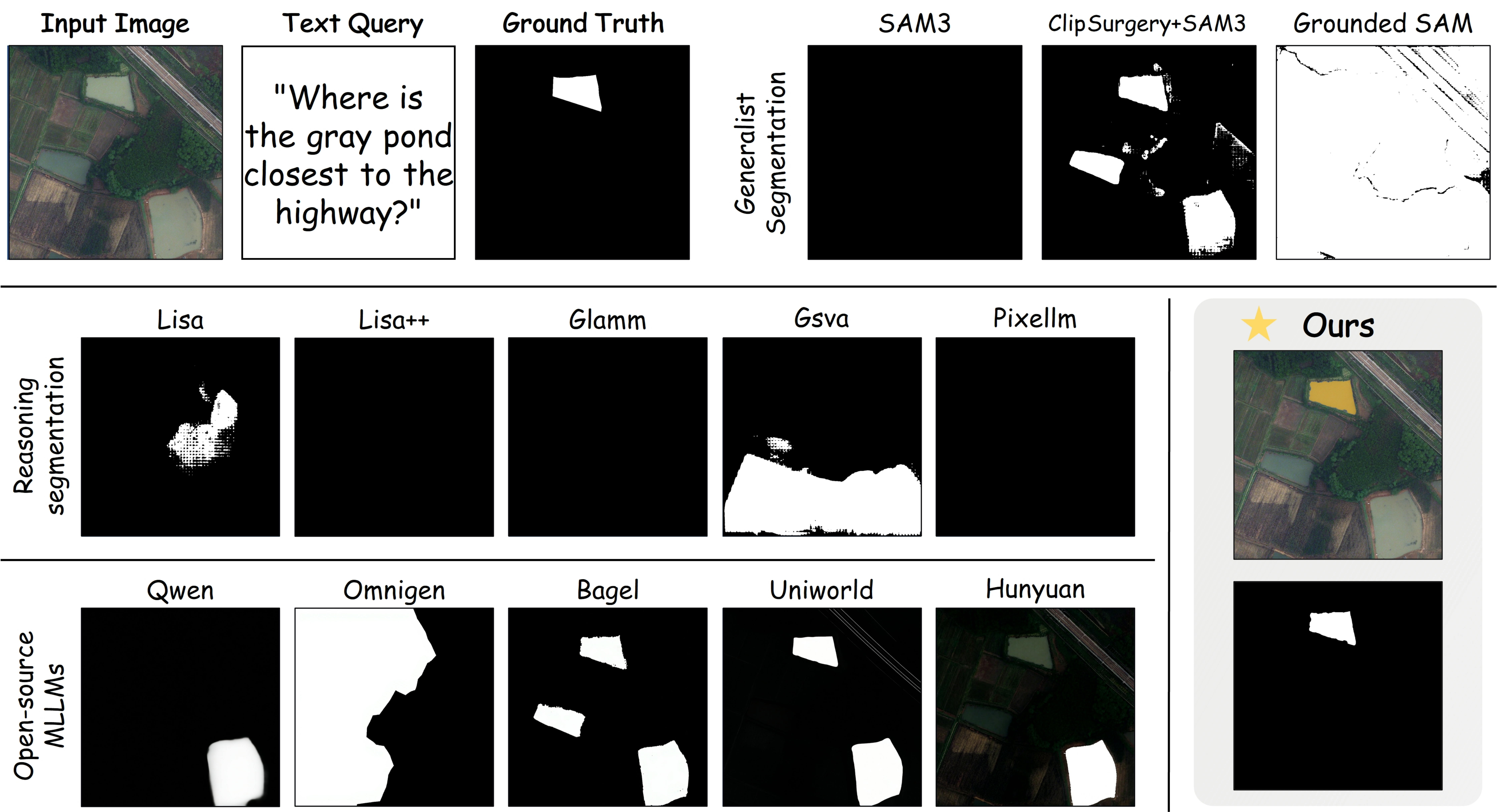}
        \caption{Case 3: Implicit instruction following.}
    \end{subfigure}
    \vspace{14pt}
    
    \begin{subfigure}{0.98\textwidth}
        \includegraphics[width=\textwidth]{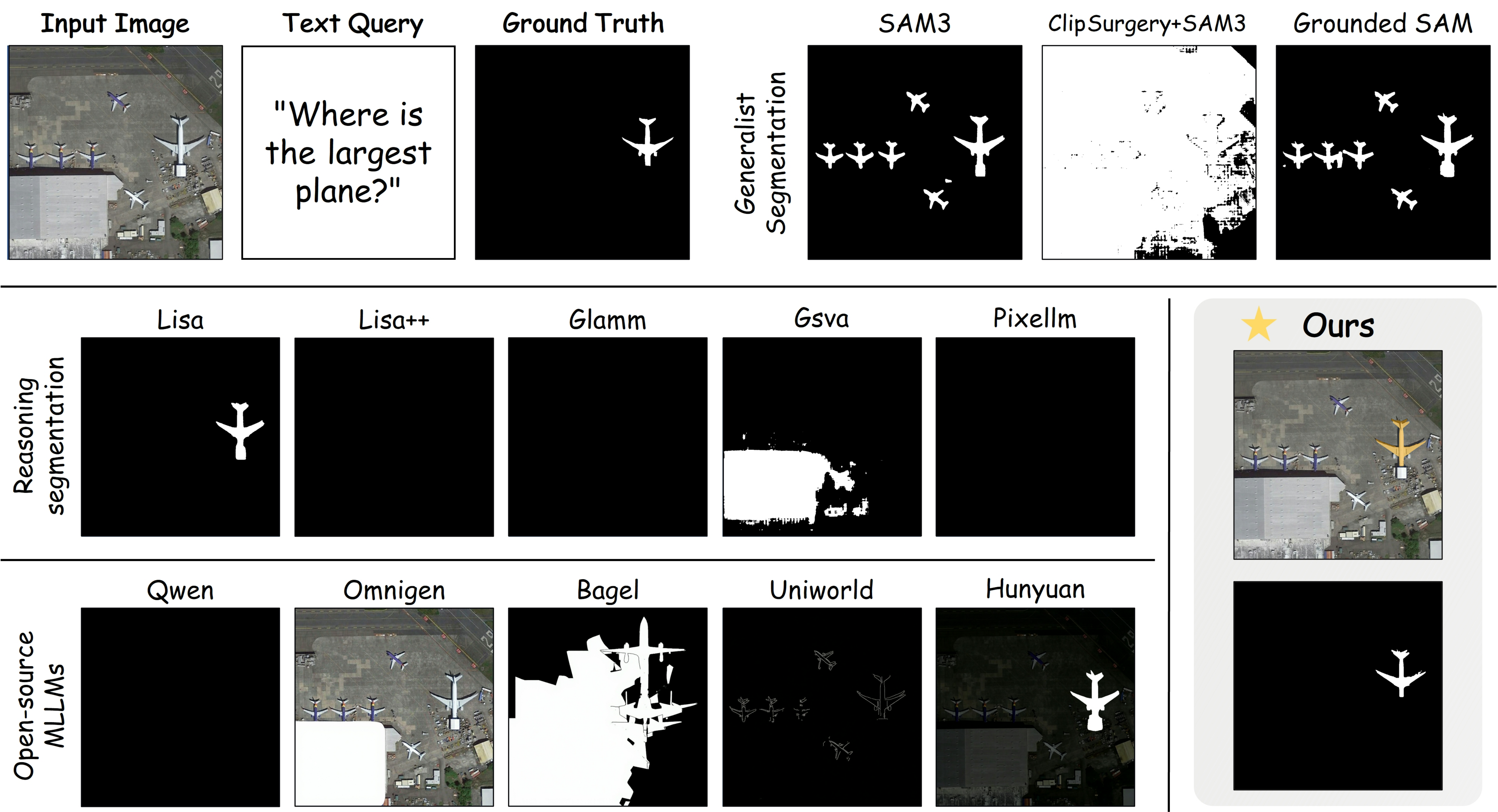}
        \caption{Case 4: Multiple target grounding.}
    \end{subfigure}
    \vspace{14pt}
     
    \caption{\textbf{Extended qualitative comparison with multiple baselines.} (Continued from previous page.)}
\end{figure}

\begin{figure}[htbp]
    \ContinuedFloat 
    \centering
    \begin{subfigure}{0.98\textwidth}
        \includegraphics[width=\textwidth]{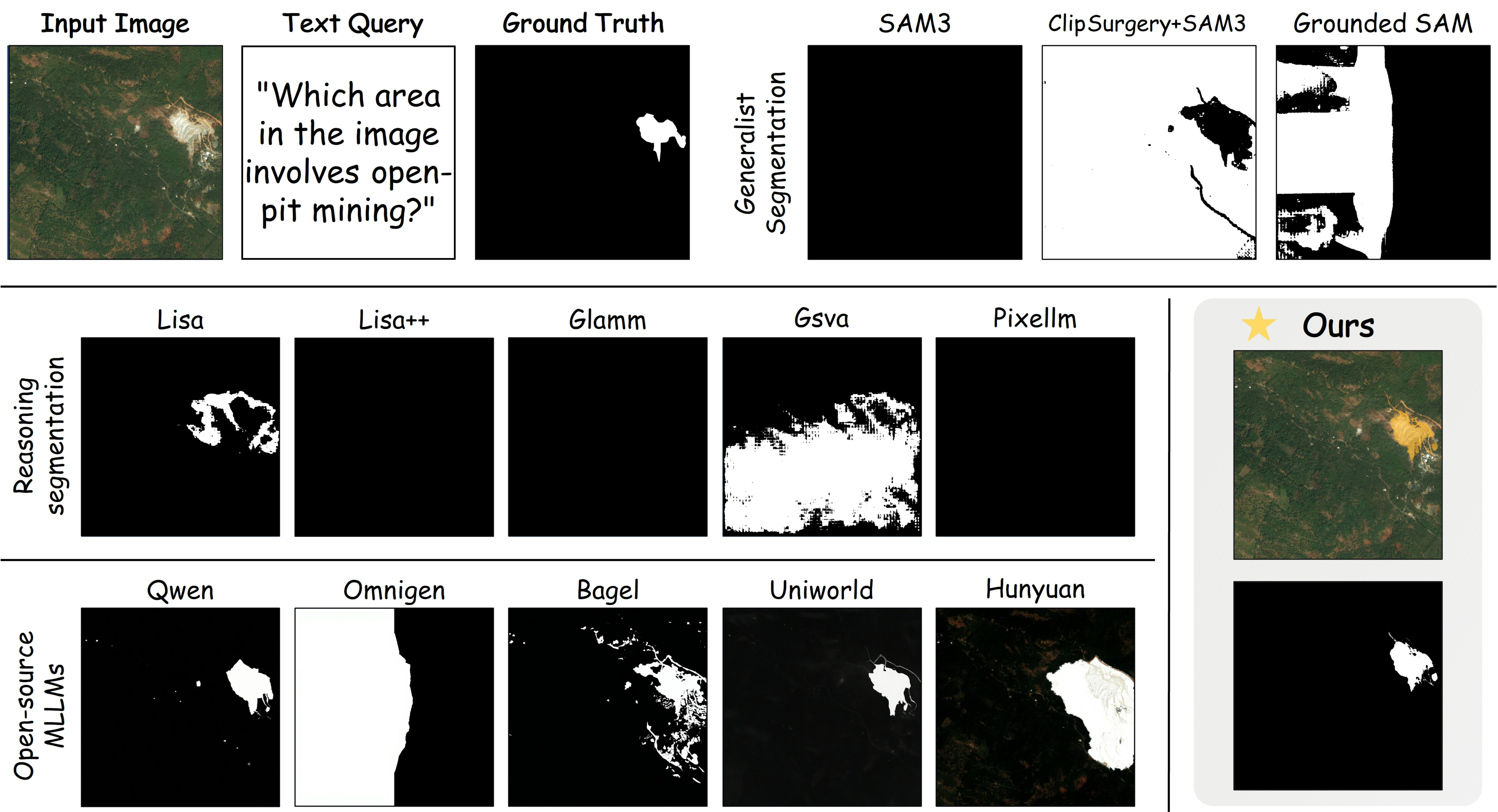}
        \caption{Case 5: Resolving multi-target spatial relationships.}
    \end{subfigure}
    \vspace{14pt}
    
    \begin{subfigure}{0.98\textwidth}
        \includegraphics[width=\textwidth]{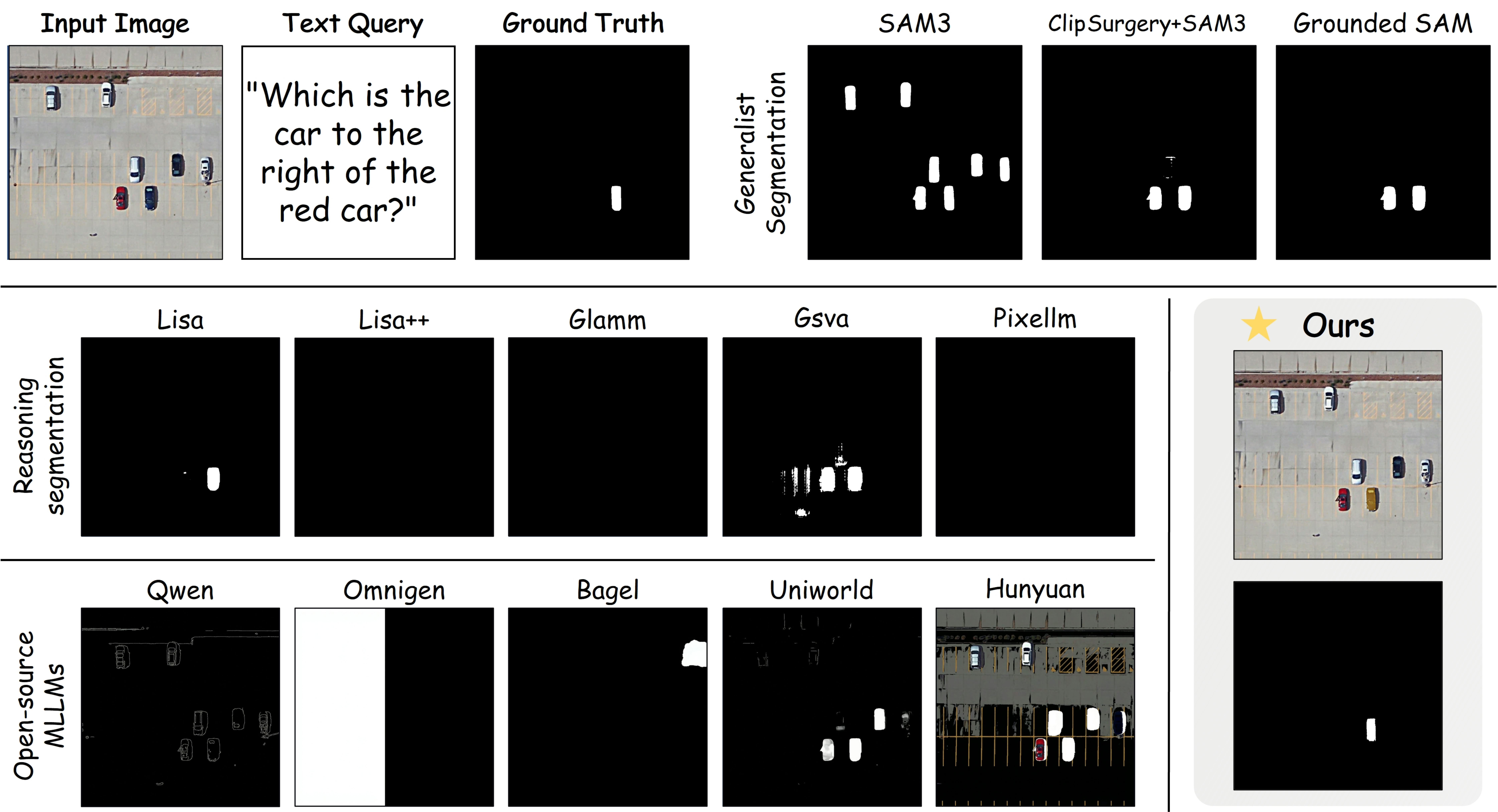}
        \caption{Case 6: Precise boundary delineation under diverse conditions.}
    \end{subfigure}

    \vspace{14pt}
    
    \caption{\textbf{Extended qualitative comparison with multiple baselines.} (Continued from previous page.)}
\end{figure}

\section{Detailed User Study Protocols}
\label{uers_study}

To comprehensively evaluate the perceptual quality and alignment of our model with human expectations, we designed a rigorous user study. We recruited 50 participants, comprising both researchers with backgrounds in computer vision and remote sensing, as well as general users, to ensure a balanced and objective assessment. 

\vspace{4pt}
\noindent\textbf{Data Preparation and Interface.}
We randomly sampled 40 challenging image-query pairs from our test set. These samples were specifically curated to include highly ambiguous queries and scenes with complex, same-class distractors. We developed a custom, web-based blind evaluation interface. For each trial, participants were shown the original image and the text query, followed by the segmentation masks generated by GeoSeg and other baseline models. To prevent any subjective bias, the order of the models was strictly anonymized and randomized for every image.

\vspace{4pt}
\noindent\textbf{Evaluation Metrics.}
Participants were instructed to evaluate the predictions using a 1-to-5 Likert scale (where 1 indicates completely unacceptable and 5 indicates perfect prediction) across three carefully defined dimensions:

\begin{itemize}
    \item \textbf{Faithfulness:} The degree to which the predicted mask accurately captures the core semantics and intent of the given text query.
    \item \textbf{Localization:} The spatial precision of the segmentation boundaries and how tightly the mask wraps the intended target without bleeding into the background.
    \item \textbf{Robustness:} The model's capability to accurately resolve linguistic ambiguities and strictly ignore surrounding same-class distractors. 
\end{itemize}

Prior to the formal evaluation, all participants completed a brief tutorial with standardized examples to ensure a consistent understanding of the scoring criteria. Furthermore, qualitative feedback was collected at the end of the survey, where evaluators explicitly highlighted the superior discriminative capability of our Dual-Route design in complex scenes.

\section{Additional Qualitative Results for Ablation Study}
\label{app:ablation} 

In this section, we present three additional qualitative examples to further illustrate the critical role of each proposed component, specifically the Bias-Aware Coordinate Refinement (\textit{Box Refine}) and the Dual-Route strategy (\textit{Route A} for Visual Cues and \textit{Route B} for Semantic Cues). 

\begin{figure}[htbp]
    \centering

    \vspace{-1mm}
    \begin{subfigure}{0.85\textwidth}
        \includegraphics[width=\textwidth]{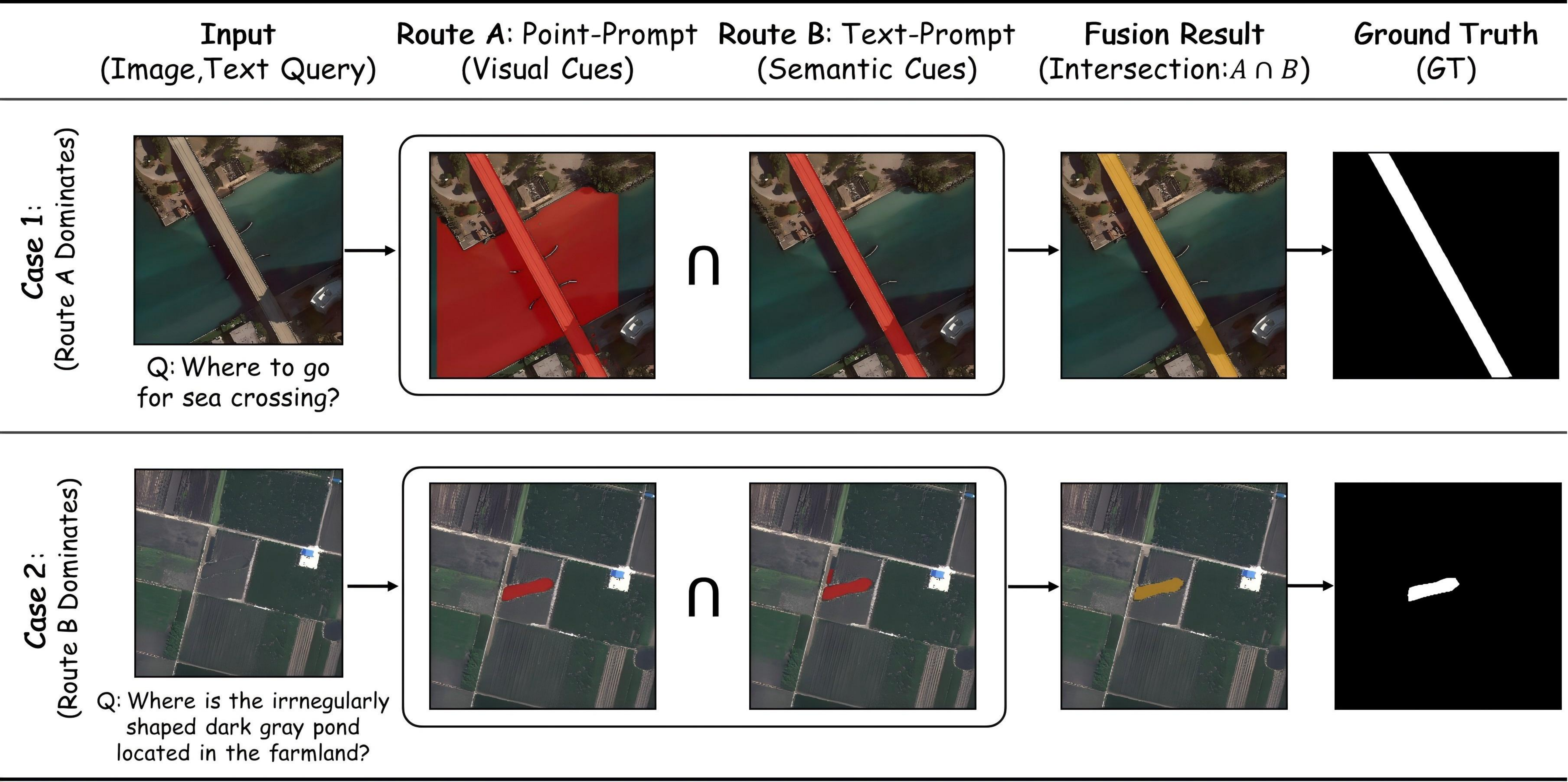}
        \caption{Case 1: Impact of missing Visual Cues (\textit{Route A}).}
    \end{subfigure}
    \vspace{5pt}
    
    \begin{subfigure}{0.85\textwidth}
        \includegraphics[width=\textwidth]{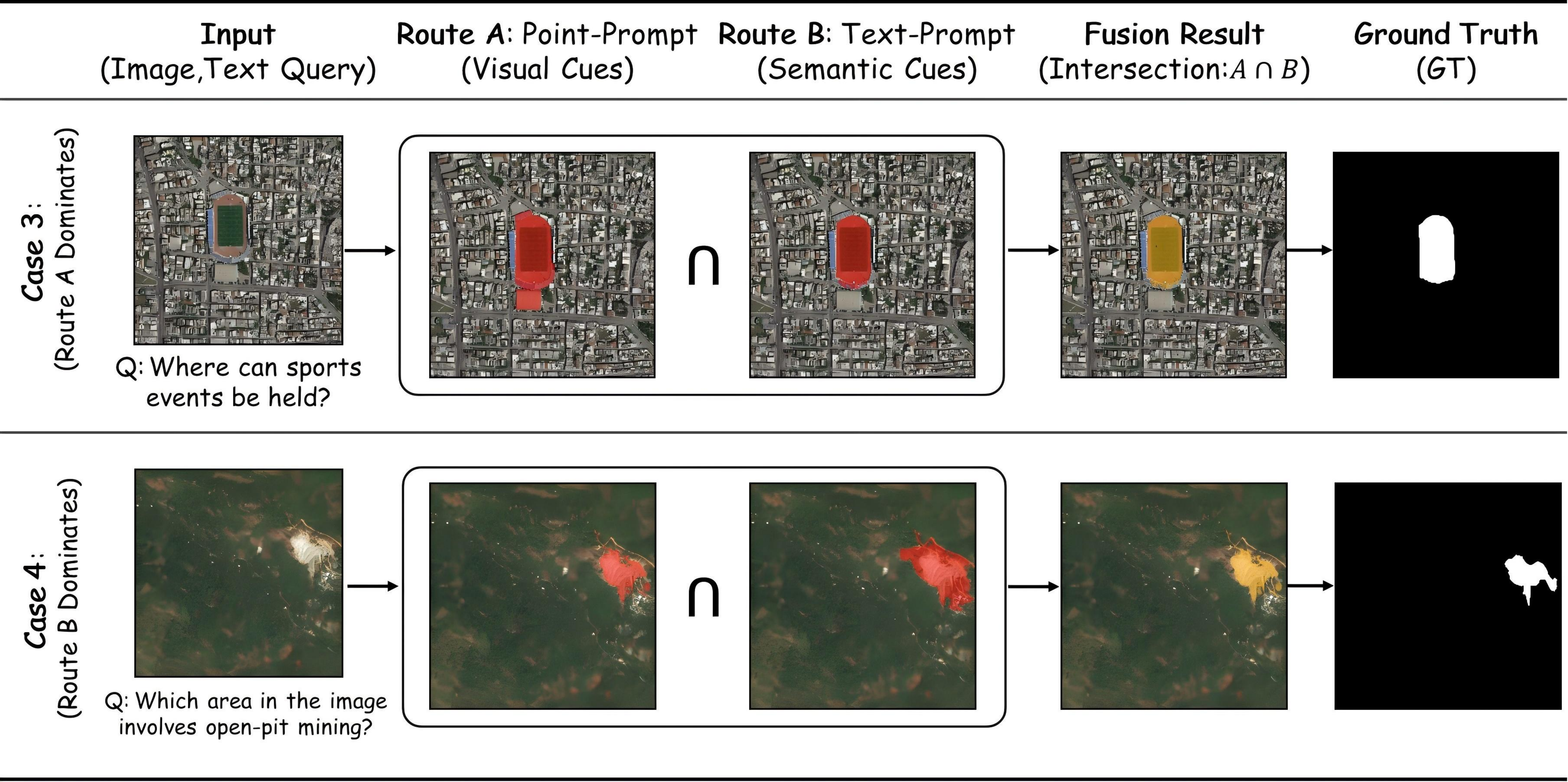}
        \caption{Case 2: Impact of missing Semantic Cues (\textit{Route B}).}
    \end{subfigure}
    \vspace{5pt}
    
    \begin{subfigure}{0.85\textwidth}
        \includegraphics[width=\textwidth]{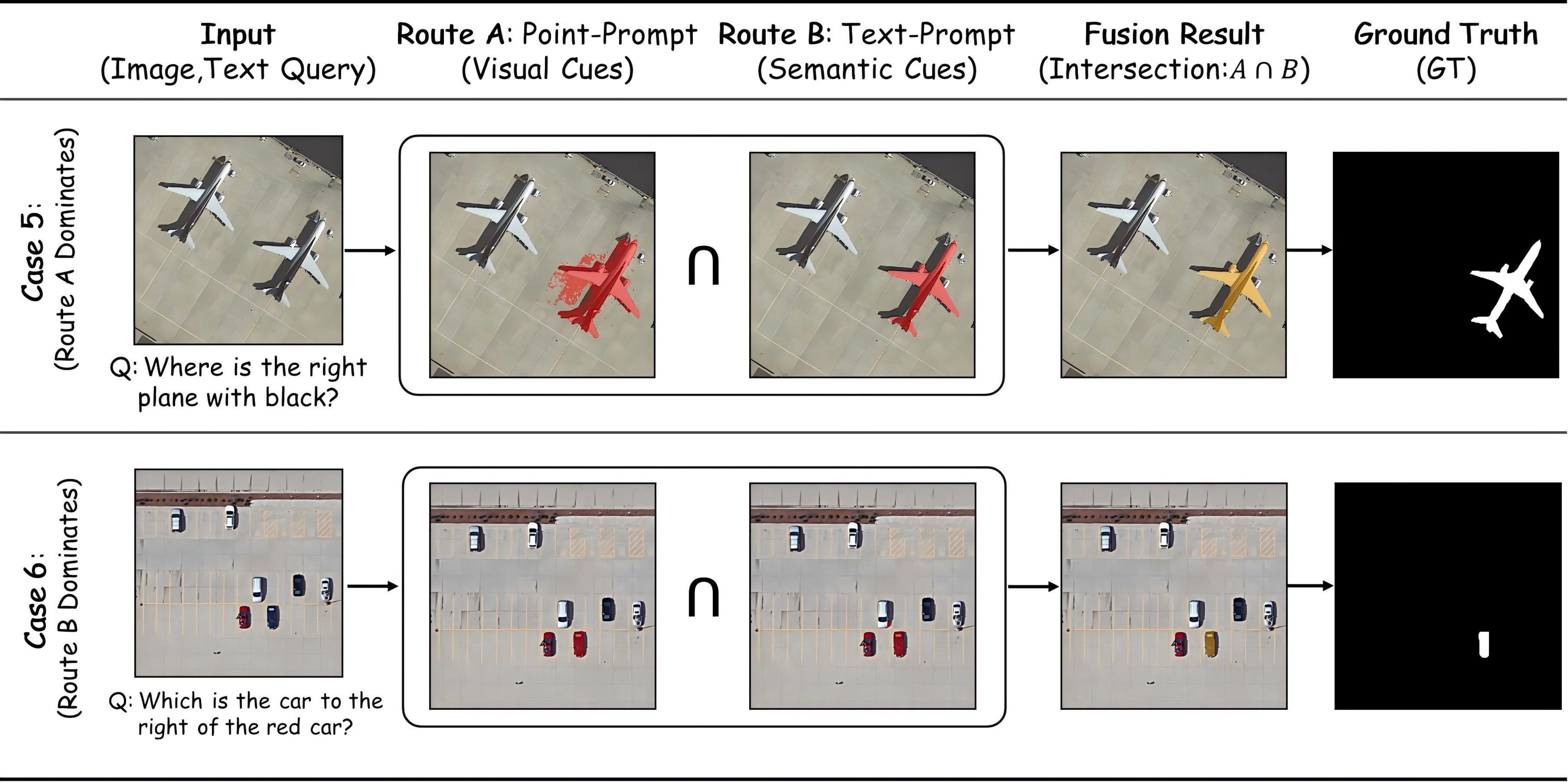}
        \caption{Case 3: Inaccurate localization without \textit{Box Refine}.}
    \end{subfigure}

    \vspace{-4pt}
    \caption{\textbf{Extended qualitative ablation study.} We showcase three additional cases comparing the full pipeline with its variants. Removing \textit{Route A}, \textit{Route B}, or the \textit{Box Refine} module results in noticeable performance degradation, such as semantic misunderstanding or coarse boundaries, verifying the necessity of our complete architecture.}
    \label{fig:more_ablation}
\end{figure}

As shown in Figure~\ref{fig:more_ablation}, missing any of these key modules leads to suboptimal predictions. Without the \textit{Box Refine} module, the model struggles with precise localization, often generating deviated boundaries. Furthermore, relying solely on either \textit{Route A} or \textit{Route B} is insufficient for comprehensive understanding, resulting in incomplete masks or incorrect grounding. In contrast, our full pipeline seamlessly integrates these complementary modules, consistently achieving the most accurate segmentation masks across all three cases.

\section{Overview of the Benchmark Dataset}
\label{app:benchmark_dataset}

In the evaluation phase of this study, we utilize 10,000 publicly available training instances from the LaSeRS dataset, originally proposed for SegEarth-R2, as our independent benchmark. Since the official test set of LaSeRS remains unreleased, these 10,000 instances serve to comprehensively assess model performance across diverse scenarios. The data organization, task dimensions, and instance categories of this benchmark are outlined below.

\vspace{4pt}
\noindent\textbf{Data Organization.}
Each instance in this benchmark is structured as a question-answer-mask triple. Alongside precise segmentation masks for pixel-level evaluation, the dataset provides bounding boxes for coarse localization, as well as the corresponding textual instructions and standard textual answers.

\vspace{4pt}
\noindent\textbf{Core Task Dimensions.}
The benchmark systematically covers four critical dimensions of language-guided segmentation in remote sensing imagery:
\begin{itemize}
    \item \textbf{Hierarchical Granularity:} The tasks encompass macroscopic semantic-level queries (e.g., ``all tennis courts''), instance-level queries (e.g., ``the rightmost tennis court''), and microscopic part-level queries (e.g., ``the service area of the tennis court'').
    \item \textbf{Target Multiplicity:} The dataset includes both single-target instructions and complex queries demanding the simultaneous grounding of multiple distinct targets.
    \item \textbf{Reasoning Requirements:} Queries range from explicit descriptions of visual attributes to implicit instructions that require the deduction of target regions via geographic commonsense (e.g., inferring a safe refuge during an earthquake).
    \item \textbf{Linguistic Variability:} The textual instructions feature significant linguistic diversity, spanning from concise short queries to highly detailed, long descriptions.
\end{itemize}

\vspace{4pt}
\noindent\textbf{Instance Categories and Scope.}
Covering a broad spectrum of remote sensing scenarios, the benchmark comprises 122 object categories. The highlighted instances are distributed across three distinct conceptual levels:
\begin{itemize}
    \item \textbf{General Categories:} Common geographical features and facilities, such as airplanes, buildings, bridges, harbors, and various sports fields.
    \item \textbf{Fine-grained Concepts:} Specialized objects differentiated by specific models or functions, including specific passenger jets (e.g., Boeing 737, A350), cargo trucks, container cranes, and engineering ships.
    \item \textbf{Part-level Elements:} Focuses on the precise segmentation of minute details under extreme scale variations, such as airplane engines, the bow or stern of a ship, football nets, and tennis court service boxes.
\end{itemize}

\end{document}